\pdfoutput=1
\documentclass[11pt]{article}

\usepackage[T1]{fontenc}
\usepackage[utf8]{inputenc}
\usepackage{mathptmx}          
\usepackage{courier}           
\usepackage[letterpaper,margin=1in]{geometry}
\usepackage{amsmath,amssymb,amsfonts,mathtools}
\usepackage{graphicx}
\usepackage{array}
\usepackage{booktabs}
\usepackage[font=small,labelfont=bf]{caption}
\usepackage{microtype}
\usepackage[numbers,sort&compress]{natbib}
\usepackage{enumitem}
\usepackage{xcolor}
\usepackage[colorlinks=true,linkcolor=black,citecolor=black,urlcolor=blue]{hyperref}
\hypersetup{
  pdftitle={Anatomy-Informed Neural Networks: Encoding Anatomic Priors in Loss and Architecture, with an SE(3) Formulation of Guidewire-Induced Aortoiliac Deformation},
  pdfauthor={David P. Stonko},
  pdfsubject={Anatomy-informed deep learning; Lie-group vascular mechanics},
  pdfkeywords={anatomy-informed neural networks; physics-informed learning; endovascular robotics; robotic navigation; EVAR; TEVAR; vascular deformation; optimal transport; SE(3); Lie algebra; Wasserstein distance; obstacle problem; variational inequality}
}

\newcommand{\SE}{\mathrm{SE}}
\newcommand{\SO}{\mathrm{SO}}
\newcommand{\se}{\mathfrak{se}}

\newcommand{\R}{\mathbb{R}}
\newcommand{\Exp}{\operatorname{exp}}
\newcommand{\Log}{\operatorname{log}}
\newcommand{\Lcal}{\mathcal{L}}
\newcommand{\Dcal}{\mathcal{D}}

\newcommand{\Rlum}{R_{\mathrm{lumen}}}
\newcommand{\State}[1]{State~#1}

\title{\bfseries Anatomy-Informed Neural Networks:\\
Encoding Anatomic Priors in Loss and Architecture,\\
with an SE(3) Formulation of Guidewire-Induced Aortoiliac Deformation}

\author{David P. Stonko, MD, MS\thanks{Division of Vascular Surgery and Endovascular Therapy,
Department of Surgery, The Johns Hopkins Hospital, Baltimore, MD, USA.
Correspondence: \texttt{dstonko1@jhmi.edu}. ORCID: 0000-0002-2804-2857.}}

\date{}

\begin{document}
\maketitle

\begin{abstract}
\noindent
Deep-learning models of anatomy can be numerically plausible yet anatomically impossible, and they
generalize poorly when data are scarce. We introduce Anatomy-Informed Neural Networks (AINN), in
which soft anatomic priors enter as penalty terms in the loss (e.g., a branching penalty that treats
a renal transplant artery off the iliac instead of the aorta as unexpected rather than impossible), in
direct analogy to a physics-informed neural network, and hard anatomic priors (e.g., continuity of
the vessel) are built into the architecture and state representation, making such invalid
predictions impossible by construction wherever the prior admits architectural enforcement.

\smallskip\noindent
We develop it on a clinical test case with limited data: how the aortoiliac tree deforms when a
stiff wire is introduced endoluminally. This is important to contemporary aortic surgery and will matter to
autonomous endovascular navigation. We lift the vessel centerline and the wire path from $\R^3$ to
curves of frames in the Lie group $\SE(3)$, and couple a Cosserat-rod wire to a tortuosity-modulated, anatomically
anchored vessel through a unilateral lumen-contact inequality. The prediction is a constrained
minimizer of the coupled elastic energy, with contact forces as its Lagrange multipliers.
Supervision is a Wasserstein-2 optimal-transport loss between the predicted projection through the
C-arm geometry and the observed angiogram, so a 2D angiogram can train a 3D prediction. The
kinematics, loss and projection are verified against known ground truth; the mechanics solver only
against its own optimality conditions, and predicted displacement is not yet mesh-converged. Here,
no network is trained. Future work will transfer this in silico model to real CT scans and test
whether it improves predictive accuracy and reduces the training data required.

\end{abstract}

\section{Introduction}
\label{sec:intro}

Deep learning dominates medical image analysis, yet two limitations persist in surgical and
interventional settings. Models that segment, register, or predict anatomy can produce outputs
that are numerically plausible but anatomically impossible
\citep{oktay2018acnn,byrne2023topological}, such as a vessel branching from the wrong parent or
a structure leaving the body. Annotated datasets for a given procedure are also small
\citep{xie2021domainknowledge}, so a network trained from scratch on a few dozen cases generalizes
poorly. Both failures arise because the model must learn anatomic structure implicitly from data
rather than being told the structure it must respect.

Physics-informed neural networks (PINNs) \citep{raissi2019pinn,karniadakis2021piml} address the
analogous problem in the physical sciences by writing a system's governing equations directly
into the training loss, rewarding the network for obeying known physics rather than
rediscovering it. We propose the anatomic counterpart. An Anatomy-Informed Neural Network (AINN)
encodes prior knowledge of anatomic structure into the model (Figure~\ref{fig:concept}), both
softly through penalty terms in the loss, in direct analogy to the physics residuals of a PINN,
and as hard constraints built into the architecture and state representation, so that whole
classes of anatomically invalid prediction are impossible by construction rather than merely
discouraged.
These priors span topology, mobility, symmetry, shape, atlas-relative position, and contact, and
AINN provides a common framework for assembling them. The soft route matters because anatomy
varies. Priors learned from a population that includes rare variants treat an unusual
configuration as improbable rather than forbidden, so an autonomous system is not caught off
guard by anatomy it has effectively already seen (\S\ref{sec:hardsoft}).

We develop the paradigm on a concrete, clinically consequential problem in endovascular surgery.
Aortic stent grafts are planned on a preoperative computed tomography (CT) angiogram treated as a
fixed map of the patient's anatomy. The vascular tree, however, is compliant, and when a stiff
guidewire is advanced the aorta straightens, the iliac arteries foreshorten, and branch-vessel
ostia migrate \citep{kaladji2013prediction,vankeulen2010neck,emendi2023guidewire}. The
preoperative configuration, which we call \State{1}, is thus replaced intraoperatively by a
deformed configuration, \State{2}, in which the device is deployed. Each subsequent maneuver
produces a further state, \State{3}, \State{4}, and so on, each predictable from the preceding one
given prior knowledge of how that maneuver reshapes the surrounding anatomy. This geometric
infidelity can compromise sealing-zone assessment and branch-vessel cannulation, and it becomes a
fundamental obstacle as procedures move toward semi-autonomous robotic navigation, where a
controller guided by a \State{1} map aims at anatomy that no longer exists at that location.

Predicting \State{2} from \State{1} is a demanding test of the AINN idea. The wire is not the
deformed vessel centerline. Because its radius is much smaller than the lumen, it rides the inner
curvature and takes the chord across bends, so models that assume wire and vessel center coincide
are structurally wrong. The vessel's resistance to deformation is heterogeneous and anatomically
anchored, the intraoperative ground truth is typically a single two-dimensional angiogram rather
than a second CT, and a realistic single-center cohort numbers only a few dozen paired cases. A
useful predictor must therefore respect the geometry of a thin stiff structure in a curved
compliant tube, learn from two-dimensional supervision, and remain data-efficient.

This problem has been studied before using finite-element methods that mesh the arterial wall \citep{kaladji2013prediction,gindre2017guidewire,mohammadi2018planning,%
emendi2023guidewire,mozahem2025renal,avril2021review}. We take the centerline-with-frames as the primitive object rather than as a quantity derived from a mesh (\S\ref{sec:worked}), which places the state space in the Lie group $\SE(3)$.

Our instantiation combines four components in a configuration we have not found applied to this problem. Every vessel and wire frame is an element of $\SE(3)$,
which separates curves that are close in space but differently oriented and guarantees valid
rigid-body configurations. A Cosserat-rod wire is coupled to a tortuosity-modulated, anatomically
anchored vessel through a unilateral lumen-contact constraint, so the equilibrium reproduces the
clinically observed bowstring deformation and apex-concentrated wall loading. Supervision uses a Wasserstein-2 optimal-transport loss between measures of frames, pushed through the C-arm projection so a single view can supervise a three-dimensional prediction. An $\SE(3)$-equivariant
residual network composed with the physics model realizes the map, so the physics does most of the
work and the network learns only a residual correction.

None of the four is new on its own \citep{murray1994robotic,bergou2008der,tang2012cosserat,duits2018reedsshepp,hu2019topology,byrne2023topological,bon2025otse2,breininger2019stiffwires,zhou2018instantiation}. \S\ref{sec:discussion} sets out what we do and do not claim.

This paper makes four contributions. It defines AINN as a methodological paradigm with an explicit parallel to physics-informed learning, and organizes anatomic priors into categories that make the hard-versus-soft design decision explicit, identifying the mobility and anchoring prior as the category we could not find in existing taxonomies (\S\ref{sec:paradigm}). It formulates
stiff-wire vascular deformation as a coupled variational problem on $\SE(3)$ with a unilateral
lumen contact, a tortuosity-anchored stiffness field, and an optimal-transport loss
(\S\ref{sec:kinematics}--\S\ref{sec:architecture}). It reports a numerical verification of these components on synthetic problems with known ground truth, including a wire-stiffness sensitivity
analysis that identifies which conclusions are robust and which are not
(\S\ref{sec:numerical}). And it sets out what separates the verified framework from clinical deployment (\S\ref{sec:discussion}).

\section{Anatomy-Informed Neural Networks: A Methodological Paradigm}
\label{sec:paradigm}

\subsection{The PINN precursor}
\label{sec:pinn}

The defining element of a physics-informed neural network \citep{raissi2019pinn} is the inclusion
in its loss function of penalty terms that enforce known physics, such as residuals of governing
partial differential equations, conservation laws, and boundary and initial conditions. A PINN
$f_\theta$ is trained to minimize a composite loss
\begin{equation}
\Lcal_{\mathrm{PINN}}(\theta) \;=\; \lambda_{\mathrm{phys}}\,\Lcal_{\mathrm{physics}}(f_\theta)
\;+\; \lambda_{\mathrm{data}}\,\Lcal_{\mathrm{data}}(f_\theta;\Dcal),
\end{equation}
where $\Lcal_{\mathrm{physics}}$ measures how badly $f_\theta$ violates the governing PDEs,
typically $\sum_i \lVert \mathcal{N}[f_\theta](x_i)\rVert^2$ for a PDE operator $\mathcal{N}$
evaluated at collocation points $x_i$, and $\Lcal_{\mathrm{data}}$ measures fit to observed data.
The structural insight is that known physics need not be learned implicitly from data, and can
instead be enforced explicitly through the loss. PINNs therefore generalize better with less data and are interpretable in physical terms. The physics is enforced through the objective, so it is
approached in the limit of successful optimization rather than guaranteed.

We propose a similar strategy for anatomy. Define an \emph{Anatomy-Informed Neural Network} as a
deep learning model whose loss function is
\begin{equation}
\Lcal_{\mathrm{AINN}}(\theta) \;=\; \sum_k \lambda_k\,\Lcal^{(k)}_{\mathrm{anatomy}}(f_\theta)
\;+\; \lambda_{\mathrm{data}}\,\Lcal_{\mathrm{data}}(f_\theta;\Dcal),
\label{eq:ainn}
\end{equation}
where each $\Lcal^{(k)}_{\mathrm{anatomy}}$ enforces one category of \emph{anatomic structure},
not the dynamics of a biological process, which is already well described as a BINN
(\S\ref{sec:relatedwork}), but the static or quasi-static structural priors that medical images and
surgical anatomy obey. The intended consequences are the same in anatomic terms, namely better generalization from less data and interpretability. We state them as design goals of the paradigm rather than as demonstrated results, since no AINN is trained in
this work. The same reasoning suggests it may mitigate anatomic hallucination, meaning the network producing
outputs that look like anatomy but cannot exist, such as a vessel that branches off the wrong
parent or a structure outside the body.

\subsection{Hard versus soft anatomic priors}
\label{sec:hardsoft}

\begin{figure}[t]
\centering
\includegraphics[width=\textwidth]{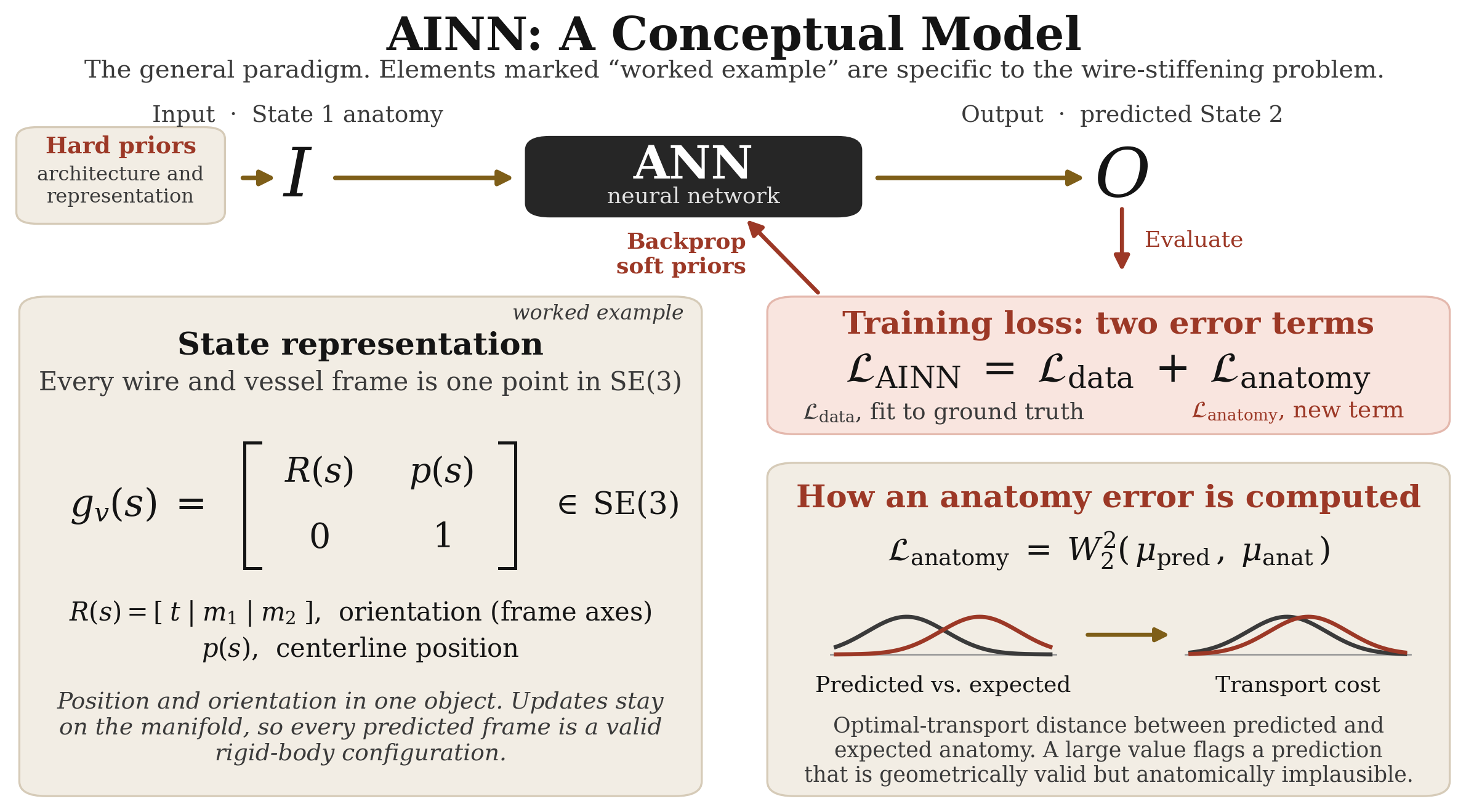}
\caption{\textbf{The AINN conceptual model.} A neural network maps an input anatomy (I, preoperative \State{1}) to an output prediction (O, intraoperative \State{2}), and anatomy enters in two ways. Hard priors are built into the representation and architecture, so every frame is a point in $\SE(3)$ and every predicted frame is a valid rigid-body configuration. That per-frame guarantee does not by itself exclude every anatomically invalid configuration, as \S\ref{sec:kinematics} sets out. Soft priors enter through the training loss, which adds an anatomy error to the data-fit error, and the lower right panel shows the general form such an anatomy term can take. In the vascular worked example the anatomic content is carried almost entirely by hard priors and the soft $\Lcal_{\mathrm{anatomy}}$ slot is specified but not exercised, the Wasserstein-2 machinery there serving the data term that compares predicted projection to observed angiogram. ANN, artificial neural network. $\SE(3)$, special Euclidean group.}
\label{fig:concept}
\end{figure}

Within the AINN framework an anatomic prior can be enforced in two complementary ways
(Figure~\ref{fig:concept}). A \textbf{hard prior} is enforced through the model's
\emph{architecture}, so the model is structurally incapable of producing predictions that violate
it, regardless of weights. Examples appearing in the worked example below include predicting
frames in $\SE(3)$, so that every prediction is a valid rigid-body configuration and the model
cannot output an off-manifold frame; composition via $\hat g = \mathrm{PhysicsSim}\cdot
\Exp(\hat\eta_\theta)$, where the matrix exponential keeps any update on the manifold; an
$\SE(3)$-equivariant network architecture; and a Cosserat-rod simulator carrying a hard
inequality on lumen contact, so the simulator cannot output interpenetrating wire-vessel
configurations. The reconstruction $g_{i+1}=g_i\Exp(\Delta s_i\hat\xi_i)$ additionally makes the
assembled centerline connected by construction, so a discontinuous vessel is not representable
(\S\ref{sec:kinematics}). What is \emph{not} conferred by the exponential map is validity relative to surrounding anatomy, meaning branching topology, absence of self-intersection, and containment within the body. \S\ref{sec:kinematics} sets out which of the three the representation delivers, \S\ref{sec:envelope} how containment is promoted to a hard constraint of the same unilateral form, and \S\ref{sec:discussion} which of them the present work actually implements.

A second reason to use $\SE(3)$ specifically, beyond its geometric role in the worked example, is
forward compatibility with downstream models. Predicting \State{2} in $\SE(3)$ means the output
lives in the same mathematical space that any subsequent endovascular-navigation model will need,
since every navigation step is a rigid-body transformation. By choosing $\SE(3)$ as the \State{2}
representation, AINN produces not only a prediction but the interface that downstream navigation
steps plug into.

A \textbf{soft prior} is enforced through the loss. The model is penalized for violations but is
not structurally prevented from making them. Examples include a topology penalty using the
Wasserstein distance between persistence diagrams \citep{hu2019topology,clough2022topological},
an atlas-position Mahalanobis distance, and a branching-graph edit distance.

Hard enforcement is preferable where it is mathematically achievable and anatomically certain,
for instance that a vessel does not leave the body or become discontinuous, since hard priors give
correctness guarantees and clean gradients. Soft enforcement is the right fallback when hard
enforcement is not architecturally possible, or when the prior \emph{should} not be hard because
real anatomy varies along a distribution rather than on a manifold.

A concrete case makes the second reason clear. Consider the topology prior, the rule specifying
which arteries branch from which. Enforced as a hard constraint, a fixed branching template treats
any departure from the canonical pattern as impossible. A patient with a renal transplant is
exactly such a departure, since the graft is anastomosed to the iliac artery and adds a branch
absent from the native tree. A model that hard-codes the normal topology cannot represent it, and
will either ignore the vessel or force the anatomy back onto the template, a dangerous failure for
autonomous navigation. A soft prior avoids this. A branching-pattern penalty learned from a
population that includes transplant recipients assigns such a configuration a low but nonzero
probability, so the graft artery is treated as unusual rather than forbidden. This is one reason
AINN is more naturally loss-based than PINN. Physical laws are fixed, whereas human anatomy lives
along a distribution, and the priors describing it must bend to accommodate real variation. It
also means that if the training set includes transplant recipients, the model simply learns the
configuration as another anatomic variant and stops penalizing it.

In practice an AINN's loss decomposes as a primary data term plus soft regularizers, with the bulk
of the anatomic content delivered by hard architectural choices. The vascular worked example is
firmly in this regime. Nearly all the anatomic prior, meaning the geometric state space via
$\SE(3)$, symmetry via equivariance, contact via the simulator's inequality, and mobility via the
anchoring field, is enforced architecturally, and the loss reduces essentially to a single
Wasserstein-2 data term comparing predicted projection to observed angiogram. Soft anatomy
penalties are available but optional. This mirrors how PINN works in practice, with hard priors
such as automatic differentiation and choice of architecture supplemented by a small number of
soft penalty terms.

\subsection{Categories of anatomic prior}
\label{sec:taxonomy}

We find it useful to organize anatomic priors into categories, each corresponding to a class of penalty terms that can enter Equation~\eqref{eq:ainn}. Prior knowledge in medical imaging has been surveyed repeatedly and thoroughly \citep{nosrati2016priors,xie2021domainknowledge,%
bohlender2023shape}, and the broader question of where knowledge enters a learning system has its
own taxonomy \citep{vonrueden2023informed}. In particular, \citet{xie2021domainknowledge} already enumerate anatomic priors covering shape, location and topology, and also classify
methods by whether knowledge enters at the input, the features, the architecture, the loss, or the
training procedure. What we add is smaller and more specific: one category we have not found in those taxonomies, and a framing that makes the hard-versus-soft decision an explicit per-prior design choice rather than an implicit consequence of method selection.

The categories below are \emph{examples}, not an exhaustive list, and the relevant priors depend
on the anatomic domain. A model for cerebral angiography would emphasize different priors than one
for the abdominal aorta, and a model for colonic geometry different again. Some priors enforce
accuracy at high-risk locations, for instance a mobility prior on the visceral segment protecting
the position of the SMA ostium during FEVAR, whereas others are chosen because they are expected
to reduce the training data required, such as symmetry or shape-distribution priors.

\paragraph{Topological priors.} Connectivity, branching cardinality, absence of crossings,
closure. A blood vessel is a connected tubular tree with a specific branching pattern, a colon is a
closed-loop structure, a gyrus has a characteristic topology. Concrete loss forms use
persistent-homology distances, for instance the Wasserstein distance between persistence diagrams
of the prediction and a reference \citep{hu2019topology,clough2022topological,byrne2023topological}.

\paragraph{Mobility and anchoring priors.} Anatomic structures are not equally free to move. The
aortic arch is anchored by the great-vessel takeoffs, the iliac arteries are mobile in the
retroperitoneum, the spine is rigid. A position-dependent stiffness field $k_{\mathrm{anat}}(x)$
encodes this, and the loss penalizes predicted displacements weighted by the local anchoring
stiffness. These parameters may be defined anatomically or inferred from CT Hounsfield units. We
have not found this category in existing taxonomies, and it is the prior that does the most work
in our example.

\paragraph{Symmetry priors.} Bilateral symmetry of paired organs, rotational invariance of
cross-sectional anatomy, $\SE(3)$-equivariance under patient pose. These are typically enforced
through the architecture using group-equivariant networks
\citep{cohen2016gcnn,weiler2021coordinate}, but can also enter the loss as equivariance-violation
penalties, $\Lcal_{\mathrm{sym}} = \mathbb{E}_{g\sim G}\lVert f_\theta(g\cdot x) - g\cdot
f_\theta(x)\rVert^2$.

\paragraph{Shape-distribution priors.} Plausible anatomic shapes occupy a low-dimensional manifold
in voxel space. The anatomically constrained neural network of \citet{oktay2018acnn} learns this
manifold with an autoencoder and uses its latent distance as a regularizer. AINN treats this as one
prior category among several, $\Lcal_{\mathrm{shape}} = -\log p_{\mathrm{shape}}(f_\theta)$ for a
learned shape distribution $p_{\mathrm{shape}}$.

\paragraph{Atlas-relative position priors.} Anatomic structures occupy expected positions in a
standardized reference frame such as Talairach, MNI, or a vertebral-body-relative frame.
Predictions far from the population position distribution are anatomically implausible. A loss
form is $\Lcal_{\mathrm{atlas}} = \lVert T_{\mathrm{atlas}}(f_\theta) -
\mu_{\mathrm{atlas}}\rVert^2_{\Sigma^{-1}}$, the squared Mahalanobis distance with $\Sigma$ the
atlas covariance.

\paragraph{Contact and exclusion priors.} Anatomic structures cannot interpenetrate, tools and
devices have hard contact constraints with vessel walls, and certain regions are anatomically
forbidden for a given pathology. Loss forms use hinge-style penalties on inequality constraints,
$\Lcal_{\mathrm{contact}} = \int \max(0, h(x))^2\,dx$ for an inequality $h(x)\le 0$. The
machinery is that of constrained-CNN losses \citep{kervadec2019constrained}, applied here to
device-tissue contact rather than to segmentation region size.

\subsection{Relationship to existing work}
\label{sec:relatedwork}

Anatomic prior knowledge already enters deep learning in many forms, including shape-distribution
regularization \citep{oktay2018acnn}, topology-aware losses
\citep{hu2019topology,byrne2023topological,clough2022topological}, inequality-constrained losses
\citep{kervadec2019constrained}, topology preserved by construction by deforming a shape prior with learned fold-free fields
\citep{wyburd2024topology}, and anatomy-informed augmentation and segmentation \citep{kovacs2023anatomyinformed,kolokolnikov2025neurofibroma,sufyan2026aaa}, where the adjective denotes anatomy entering as input, through a deformation field, engineered features, or a region prior, rather than through the objective and the hypothesis class as it does here. These have been catalogued as categories of domain knowledge
\citep{nosrati2016priors,xie2021domainknowledge,bohlender2023shape,vonrueden2023informed,%
banerjee2025pimlmia}. AINN's nearest paradigm-level neighbor is BINN
\citep{lagergren2020binn}, which extends PINN to biology by encoding governing biological dynamics
in the loss; the difference is the type of constraint, since AINN encodes static anatomic structure
and predicts a future static configuration rather than a dynamical law. A complete physiological
model could carry both.

\subsection{Roadmap}
\label{sec:roadmap}

The remainder of this paper develops the worked example, predicting how the aortic and iliac
vascular tree deforms when a stiff endovascular wire is introduced, which is the
\State{1}~$\rightarrow$~\State{2} mapping. It is a well-described problem in endovascular surgery, and the same wire-stiffening physics applies across the aortic procedures of \S\ref{sec:worked} and to any robotic endovascular navigation platform. For tractability we focus on EVAR, where the data pipeline is most accessible, with high case volume, consistent
intraoperative imaging, both common and external iliac wire courses visible on the angiogram, and a
stiff wire, typically a Lunderquist, routinely in place during landing-zone selection.

The example is hard along several AINN axes simultaneously. It requires topological, mobility,
symmetry, and contact priors, and it must eventually be supervised under realistic constraints,
with a single 2D fluoroscopic angiogram as intraoperative ground truth paired with the same
patient's preoperative 3D CT. The components listed in \S\ref{sec:intro} map onto the prior categories above as follows: the $\SE(3)$ state space is the geometric prior, the anchored vessel and the lumen inequality are the mobility and contact priors, the equivariant residual network is the symmetry prior, and the Wasserstein-2 projected loss is the data term. Sections~\ref{sec:worked}--\ref{sec:numerical} develop the instance, and we note throughout which category each piece corresponds to, so the worked example doubles as a constructive recipe for instantiating AINN beyond EVAR.

\section{The Worked Example: Geometric Infidelity in Endovascular Surgery}
\label{sec:worked}

\subsection{The clinical problem}

The planning paradigm for thoracic and abdominal endovascular surgery, including TEVAR, EVAR,
FEVAR, BEVAR, and the increasingly common iliac-branch and parallel-graft procedures, treats the
preoperative CT angiogram as a static coordinate system. Sizing, landing-zone selection,
fenestration planning, and increasingly the trajectories of robotic delivery systems are all
calibrated against a single preoperative snapshot. In current practice the aortoiliac lumen is segmented, a centerline is traced through the aorta and both iliac systems, the vessel is
reformatted along that centerline so diameters and lengths are measured on the straightened lumen,
and those measurements are transcribed into an operative plan fixing graft component diameters,
lengths, and fixation levels. The repair is then executed under single-plane fluoroscopy.

Two features of this workflow are decisive. First, every planning measurement derives from the
preoperative CT, in which the anatomy is undeformed and no instruments are present. Second, once
the stiff wires are in place the patient is not re-imaged in three dimensions, because cone-beam
CT is not routinely acquired during EVAR. The only intraoperative record of the deformed anatomy
is a single 2D fluoroscopic angiogram, and that projection is therefore the sole ground truth
against which an intraoperative prediction can be checked. Any endovascular navigation or
roadmap-forecasting technology must accordingly express its output in, and estimate its error
against, that projection. A method whose accuracy is reported only in the 3D preoperative frame
cannot be falsified intraoperatively.

The vascular tree is not a system of rigid pipes. It is a compliant, anisotropically constrained
structure in a deformable retroperitoneum, tethered by the diaphragm and the iliac bifurcation,
supported posteriorly by the spine and laterally by retroperitoneal fat of variable density. When
a stiff guidewire such as a Lunderquist is advanced, the aorta straightens, the iliac bifurcation
migrates, and the renal and superior mesenteric ostia displace. This is the \State{1}~$\rightarrow$~\State{2} transition of \S\ref{sec:intro}.

\begin{figure}[t]
\centering
\includegraphics[width=\textwidth]{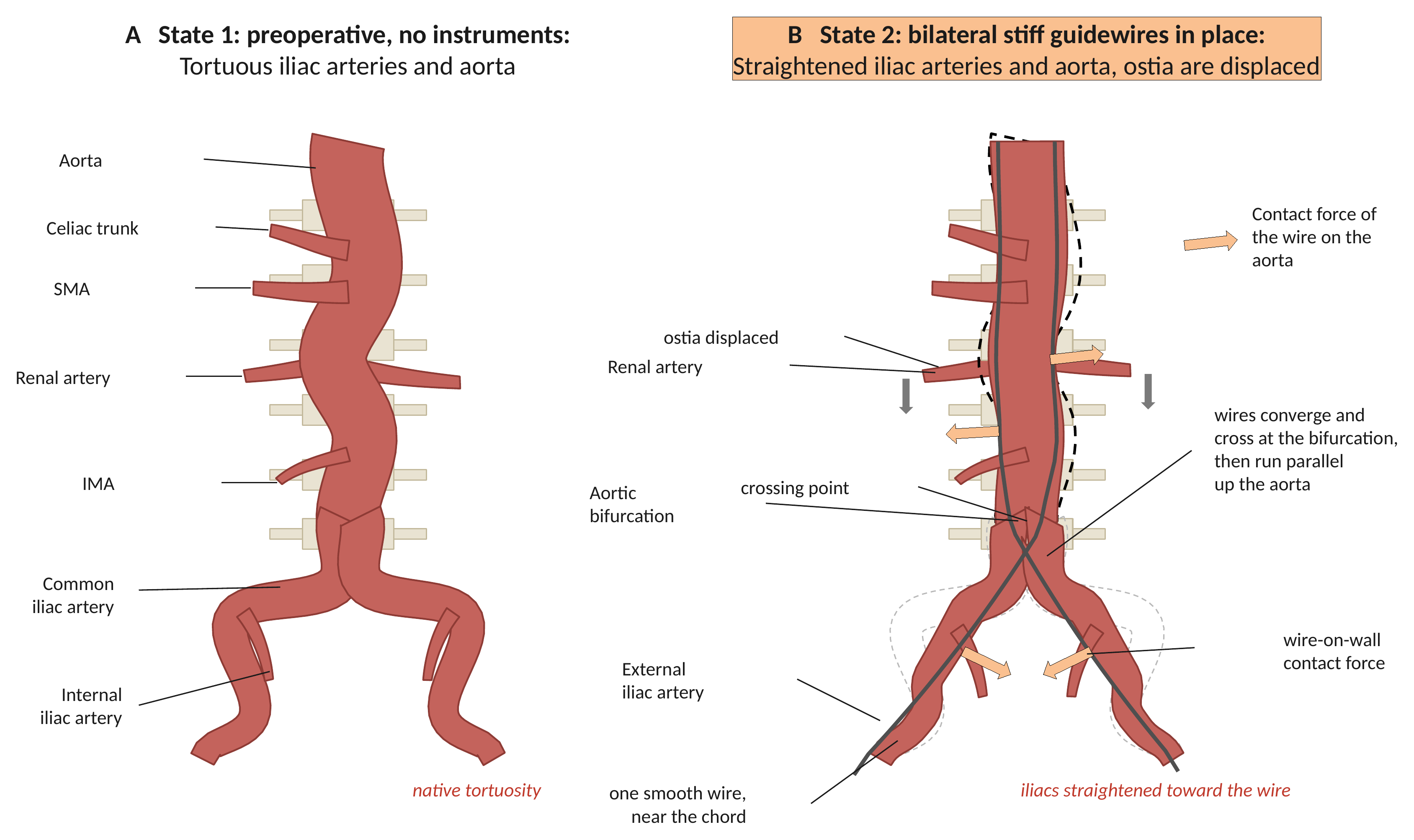}
\caption{\textbf{Aortoiliac deformation under stiff-wire loading, \State{1} $\rightarrow$ \State{2}} (schematic; not patient imaging). (A) Preoperative anatomy, with native tortuosity in the iliac arteries and the aorta and no instruments in place. (B) The intraoperative configuration after bilateral stiff guidewires are advanced from femoral access. Each wire is a single smooth, near-chord curve rather than a copy of the vessel centerline, because it is by far the stiffest member of the system. Orange arrows mark the contact force of the wire on the wall and gray arrows the resulting ostial displacement. The dashed black line traces the \State{1} aortic contour and the dashed gray outline the \State{1} iliac contour, each reproduced from panel A for comparison. The wires converge and cross at the aortic bifurcation and then run parallel up the aorta. The worked example (\S\ref{sec:kinematics}--\S\ref{sec:numerical}) models the single dominant wire and its bowstring contact (Figure~\ref{fig:bowstring}), of which panel B is the clinical superposition.}
\label{fig:aortoiliac}
\end{figure}

Two consequences follow. An aneurysm neck judged adequate on \State{1} changes its angulation in
\State{2}, and indeed \citet{vankeulen2010neck} showed that infrarenal neck angulation decreases
measurably during and after repair, so a seal zone can be misjudged. Branch-vessel ostia displace,
complicating cannulation of the SMA and renal arteries in fenestrated and branched repair, with
\citet{mozahem2025renal} reporting a mean vertical displacement of $10.4$~mm at the renal ostia
between the preoperative and intraoperative configurations in sixteen patients selected for having
experienced significant ostial displacement, and \citet{emendi2023guidewire}
reporting a $60$ to $79\%$ tortuosity reduction in the guidewire-side common iliac artery, and $21$
to $49\%$ contralaterally, under guidewire loading. In practice EVAR uses bilateral common femoral
access, so a stiff guidewire is usually present in each iliac and the two wires cross just above the
aortic bifurcation (Figure~\ref{fig:aortoiliac}). The worked example below models the single
dominant stiff wire, which is the core mechanical building block of that bilateral picture.

Guidewire-induced aortic deformation has been modeled before, by finite-element analysis on a
meshed arterial wall, with patient-specific predictions validated against intraoperative imaging
\citep{kaladji2013prediction,gindre2017guidewire,mohammadi2018planning,emendi2023guidewire,%
mozahem2025renal,avril2021review}. That work establishes that the deformation is predictable and
sets the accuracy against which any new method should be measured.

We formulate it differently, because of how the problem is actually
posed in the operating room. A finite-element treatment discretizes the arterial \emph{wall}. That
is the natural description for a continuum mechanician, and it produces
volumetric meshes for computational fluid dynamics, but it is not how the surgeon or operator reasons. A surgeon plans on the centerline, as above. Intraoperatively the operator thinks about where the wire sits, which way the catheter tip is
pointing, and how far along the centerline the next landmark lies. A robotic navigation controller
would reason about the same quantities, because every advance, rotation, and cannulation is a
rigid-body motion of a device frame relative to a vessel frame.

We therefore take the centerline and the orientation frame carried along it as the primitive
object rather than as something recovered from a mesh by post-processing. That places the state
space in the Lie group $\SE(3)$, and the rest follows: the wire and the vessel become two curves
in the same space, their coupling becomes a single inequality between them, the strain of each
becomes an element of the Lie algebra, and any predicted update is a physically realizable
rigid-body transformation by construction. It also keeps the forward model small, a few hundred degrees of freedom rather than a volumetric mesh, and supervised by the angiogram that is already acquired in every case.

\subsection{Why this is a hard test of AINN}

Predicting \State{2} from \State{1} exercises several AINN prior categories at once and is
constrained by the realities of the data. The preoperative input is a high-resolution 3D CT
angiogram, whereas the intraoperative observation is a single-view 2D fluoroscopic angiogram with
instruments in place, typically a stiff Lunderquist guidewire and a soft pigtail catheter on
separate access for contrast. The mathematics applies cleanly to both, each being a thin elastic
structure with its own bending stiffness. The stiff wire dominates the deformation and the pigtail
adds a smaller perturbation at a fraction of the stiffness within the same Cosserat-rod system \citep{tang2012cosserat,jilani2025catheter}.

Two mathematical ideas drive the approach. Lie groups provide the natural state space \citep{murray1994robotic}, with $\SE(3)$ encoding the position \emph{and} orientation of every infinitesimal vessel segment and the matrix exponential guaranteeing that any predicted deformation is a realizable rigid-body transformation. Optimal transport provides the natural loss, since the Wasserstein-2 distance measures the work required to morph one geometric distribution into another, is geometry-aware, and extends to manifold-valued measures.

\section{Kinematics on \texorpdfstring{$\SE(3)$}{SE(3)}: Two Coupled Curves}
\label{sec:kinematics}

Let $s\in[0,L]$ denote arc length along a smooth curve in the reference configuration. We model
both the vessel centerline and the guidewire path as curves of \emph{frames}, meaning points
equipped with an orientation, in the special Euclidean group
\begin{equation}
g(s) \;=\; \begin{bmatrix} R(s) & p(s) \\ \mathbf{0} & 1 \end{bmatrix} \;\in\; \SE(3),
\end{equation}
where $R(s)\in\SO(3)$ is the local rotation matrix and $p(s)\in\R^3$ the spatial position. The
lift from $\R^3$ to $\SE(3)$ separates curve passes that are close in position but differ in
orientation, which mitigates the crossing ambiguity for a vessel folded back on itself
\citep{duits2010se2,duits2018reedsshepp}. We are careful not to overclaim here. Two genuinely
adjacent vessels running near-parallel with nearly identical positions and tangents, such as the
two iliac limbs or the aorta and the inferior vena cava, remain ambiguous under the lift and are
separated by the segmentation, not by the representation.

The framework requires two such curves.
\begin{itemize}[leftmargin=1.6em,itemsep=2pt,topsep=3pt]
\item \textbf{Vessel centerline} $g_v(s)\in\SE(3)$, $s\in[0,L_v]$, obtained from the segmented
preoperative CT by attaching an adapted orthonormal frame at each centerline node. We write
$p_v(s)\in\R^3$ for the position component, meaning the actual 3D coordinates of the centerline at
arc length $s$.
\item \textbf{Guidewire path} $g_w(s)\in\SE(3)$, $s\in[0,L_w]$, the equilibrium configuration of
the inserted wire, with $p_w(s)\in\R^3$ its position.
\end{itemize}

\paragraph{Choice of frame.} The obvious choice is the Frenet-Serret frame of tangent, normal and
binormal, but Frenet frames are degenerate at near-zero-curvature points and twist erratically
along straight segments, which is precisely the regime of a nearly straight guidewire. We
therefore use a \textbf{Bishop, or parallel-transport, frame} throughout \citep{bishop1975frame},
which is stable along straight segments and gives a continuous body-fixed orientation. This choice
has a consequence that we make explicit below, because it changes what the strain components mean.

\begin{figure}[t]
\centering
\includegraphics[width=\textwidth]{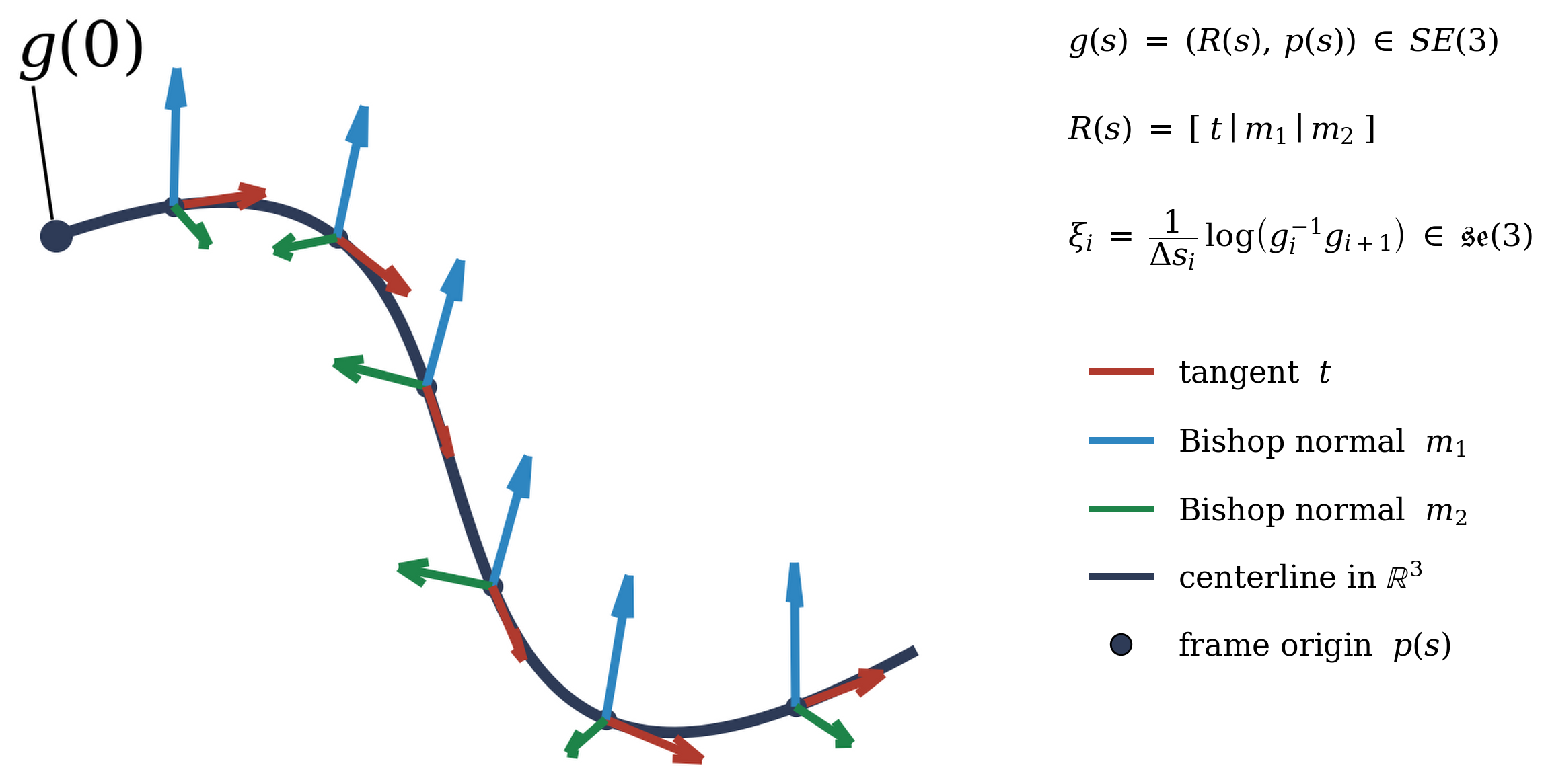}
\caption{\textbf{Lifting a vessel centerline from $\R^3$ into $\SE(3)$.} Attaching an orthonormal frame at each centerline point makes position plus orientation one element $g(s)$ of $\SE(3)$, with the incremental twist between adjacent frames and the exponential map that inverts it shown for one step.}
\label{fig:se3lift}
\end{figure}

\paragraph{Body-fixed strain.} The \emph{shape} of each curve, meaning how it bends and twists from
one node to the next, is captured by its body-fixed strain (Figure~\ref{fig:se3lift}). Walking
along the centerline one infinitesimal step at a time, at each step one translates forward along
the local tangent and rotates slightly because the curve is bending or twisting. The pair of
translation rate and rotation rate at every point is an element of the Lie algebra
$\se(3)$,
\begin{equation}
\hat\xi(s) \;=\; g^{-1}(s)\,\frac{dg}{ds} \;=\;
\begin{bmatrix} \hat\omega(s) & \nu(s) \\ \mathbf{0} & 0 \end{bmatrix} \;\in\; \se(3).
\label{eq:strain}
\end{equation}
Here $\omega(s)\in\R^3$ collects the rotational rates and $\nu(s)\in\R^3$ the translational
rates. The hat denotes the standard isomorphism embedding a $6$-vector $(\omega,\nu)$ as a $4\times4$ matrix in $\se(3)$, and $(\cdot)^\vee$ denotes its inverse, mapping a matrix in $\se(3)$ back to its $6$-vector of coordinates.

Two clarifications matter and are often glossed. First, with the tangent taken as body axis $1$,
the component $\omega_1$ is the \emph{material twist} about the wire's own axis, while
$\omega_2,\omega_3$ are the two curvature components. For a Bishop frame the defining property is
$\omega_1\equiv 0$, so the frame carries no twist by construction and the geometric torsion of
classical differential geometry is \emph{not} an entry of $\omega$. It must be recovered
separately as $\tau = d\phi/ds$, where $\phi$ is the angle of the curvature vector within the
Bishop normal plane. Second, if $s$ is arc length in the reference configuration and the rod is
inextensible and unshearable, then $\nu \equiv e_1$ carries no information. We retain $\nu$ in the
general formulation, defining $\nu = R^\top dp/ds_0$ with respect to the fixed reference arc
length $s_0$, so that $\lVert\nu\rVert$ is the stretch ratio, and we return in \S\ref{sec:energy}
to how the wire's axial degree of freedom is actually handled.

\paragraph{Discrete strain.} In numerical practice the matrix logarithm between adjacent frames
returns the \emph{incremental twist}
\begin{equation}
\hat\zeta_i \;=\; \Log\!\left(g_i^{-1}g_{i+1}\right),
\end{equation}
whose rotational part is approximately $\kappa\,\Delta s_i$ and is dimensionless. This is not
itself the strain. Dividing by the node spacing recovers the strain,
\begin{equation}
\hat\xi_i \;=\; \frac{1}{\Delta s_i}\,\Log\!\left(g_i^{-1}g_{i+1}\right),
\qquad \Delta s_i = s_{0,i+1}-s_{0,i},
\label{eq:discretestrain}
\end{equation}
where $\Delta s_i$ is the arc length of element $i$ in the reference configuration, consistent with the definition of $\nu$ above. The strain has units of inverse length in its rotational part and is the quantity that must appear in the quadratic energies of \S\ref{sec:energy}. Dividing instead by the deformed chord $\lVert p_{i+1}-p_i\rVert$, as is tempting when no reference parameterization is carried, makes the translational block of $\xi_i$ a function of the element's bend angle alone and identically of unit magnitude, so it could never report stretch. Omitting the $1/\Delta s_i$ makes the computed
bending energy scale with $\Delta s^2$ and therefore depend on the discretization. Conversely the
matrix exponential
\begin{equation}
g_{i+1} \;=\; g_i\,\Exp\!\left(\Delta s_i\,\hat\xi_i\right)
\end{equation}
advances one frame to the next. This duality, in which the Lie group is the configuration
manifold, the Lie algebra is the tangent space of strains, and the exponential map relates them,
guarantees that every integration step, and hence every predicted frame, is a valid rigid-body
transformation. It guarantees more than that. Because each frame is generated from its predecessor
by a rigid motion, a curve reconstructed from a strain sequence is \emph{connected by construction}: no strain sequence produces a gap, so a torn or discontinuous centerline is not representable at all, and the segment lengths $\Delta s_i$ are carried explicitly rather than inferred. This is the representational counterpart of an anatomic fact, that large and medium vessels do not start and stop, and it is that anatomic prior the construction encodes. Continuity of the centerline is therefore a hard prior of the representation, not a
penalty.

The exponential map does not deliver any validity that depends on where the curve sits
relative to the rest of the anatomy. Correct branching topology is imposed by the tree structure whereas non-penetration of the lumen is imposed by the contact constraint of \S\ref{sec:contact}. Absence of self-intersection is not conferred either, since a tree constrains the connectivity graph and not the geometric embedding, and would require a further unilateral constraint of the form of \S\ref{sec:envelope}; and containment within the body is not imposed structurally in
the formulation as developed here, being carried by the anchoring term $J_{\mathrm{anchor}}$ of
\S\ref{sec:Janchor}, which is a quadratic penalty and therefore a \emph{soft} prior by the
definition of \S\ref{sec:hardsoft}. A frame may be a perfectly valid element of $\SE(3)$ and sit
outside the patient. But we can fix that: \S\ref{sec:envelope} shows that an anatomic envelope enters as a second unilateral
constraint of the form already used for lumen contact, which makes containment a hard prior.
\S\ref{sec:architecture} states which of these guarantees survive composition with the learned
residual.

\section{The Unilateral Lumen Contact Constraint}
\label{sec:contact}

A central clinical observation, often elided in simplified wire-deformation models, is that
\textbf{the wire is not the deformed centerline of the vessel}. The wire's outer radius $r_w$ is
strictly less than the local lumen radius $\Rlum(s)$. As the wire minimizes its own bending
energy under axial load, it adopts a path \emph{straighter} than the centerline of the host
vessel, riding the inner curvature and taking the chord rather than the arc
(Figure~\ref{fig:bowstring}). Along the lesser curve of the aortic arch, or along the inner wall
of a tortuous external iliac, the wire bowstrings.

\subsection{Three candidate forms, and which is the constraint}

Three distinct quantities are easy to conflate. They play different roles, and only one of them is the non-penetration condition.

\paragraph{The naive same-index form.} Comparing wire and vessel at the same parameter value,
\begin{equation}
\lVert p_w(s) - p_v(s)\rVert \;\le\; \Rlum(s) - r_w,
\label{eq:naive}
\end{equation}
is adequate only when the two curves share a parameterization. Under loading they do not. When a
stiff wire is advanced into an iliac, the iliac foreshortens, so wire and vessel have different
arc lengths after deformation. Equation~\eqref{eq:naive} then conflates the \emph{radial} offset
toward the wall, which is what the constraint should detect, with \emph{axial drift} of one curve
relative to the other, which has nothing to do with hitting the wall.

\paragraph{The closest-point form, which is the constraint.} The geometrically correct
non-penetration condition for a wire of radius $r_w$ inside a tube of radius $\Rlum$ about the
centerline $\gamma_v$ is
\begin{equation}
d_c(s) \;\coloneqq\; \min_{\sigma\in[0,L_v]} \lVert p_w(s) - p_v(\sigma)\rVert
\;\le\; \Rlum(\sigma^*(s)) - r_w \qquad \forall s,
\label{eq:closestpoint}
\end{equation}
where $\sigma^*(s)$ is the achieving vessel arc length. This is the standard form in
Cosserat-rod-in-tube unilateral contact mechanics, it is parameterized by \emph{wire} arc length,
and it handles foreshortening correctly by construction. Equation~\eqref{eq:closestpoint} is the constraint we impose in both the two- and three-dimensional models of \S\ref{sec:numerical}. The right-hand side $\Rlum(\sigma^*(s))-r_w$ is the slack the wire has, and exploiting it is what produces the bowstring.

\paragraph{The in-frame offset, which is a clinical readout.} Distinct from the constraint is the
offset measured in the wire's own cross-section. At wire arc length $s$, the plane through
$p_w(s)$ orthogonal to the wire tangent $t_w(s)$ meets the vessel centerline, and the offset
within that plane, $d_{\text{in-frame}}(s)$, is the quantity that matters clinically. A stent expands circumferentially about the wire, orthogonal to the wire's tangent, so the lumen wall in the wire's frame is the wall the stent engages. Branch cannulation is targeted the same way: the operator looks down the wire and aims at where the ostium sits in that frame.

The closest-point distance and the in-frame offset coincide only when $t_w \parallel t_v$, which is
exactly the case in which there is no bowstring. We therefore keep them separate.
Equation~\eqref{eq:closestpoint} is what is enforced; $d_{\text{in-frame}}$ is what we would report
clinically, and the models of \S\ref{sec:numerical} report the constraint residual only. In three dimensions the orthogonal complement of $t_w(s)$ is a $2$-plane, and a plane meets a space curve in isolated points that need not be unique for a tortuous tree, since the normal plane at a wire node in the distal aorta may also cut the contralateral iliac limb. We select the intersection continuous in $s$ from its predecessor and restricted to the parent vessel segment. Existence and local uniqueness then follow from the implicit function theorem provided $\lvert t_w(s)\cdot t_v(\sigma^*)\rvert \ge \delta > 0$, so that the normal plane meets the centerline transversally. In two dimensions the cross-section is a line and the intersection is generically unique, so none of this arises. The step from 2D to 3D is not merely a matter of cross-sectional dimensionality.

\subsection{Unilaterality and complementarity}

The contact is \emph{unilateral}, meaning it can only push, never pull. The wire and the vessel
wall are not bonded, cannot interpenetrate, and cannot adhere. Writing the gap function
\begin{equation}
h(s) \;=\; d_c(s) - \left(\Rlum(\sigma^*(s)) - r_w\right)
\end{equation}
and the contact force magnitude $\mu(s)$, the contact conditions are the complementarity system
\begin{equation}
h(s) \le 0, \qquad \mu(s) \ge 0, \qquad h(s)\,\mu(s) = 0 .
\label{eq:complementarity}
\end{equation}
Either the wire is separated and the force is zero, or it is in contact and the force is positive,
never both at once. This contrasts with bilateral contact such as a screw thread or a weld, where
the surfaces are bonded, the constraint is an equality, and the force is sign-indefinite.

The inequality is imposed for every $s$, and it is a hard constraint in the sense of
\S\ref{sec:hardsoft}. What varies along the curve is whether it is \emph{active}. Where it is slack the wire exerts no reaction at that station, though the vessel returns to its \State{1} equilibrium there only if the constraint is inactive over a wide enough neighborhood. Where it
saturates, the wire exerts a normal reaction on the wall, and it is precisely this distribution of
reactions, concentrated, asymmetric and sign-definite, that drives the
\State{1}~$\rightarrow$~\State{2} deformation.

\begin{figure}[t]
\centering
\includegraphics[width=0.62\textwidth]{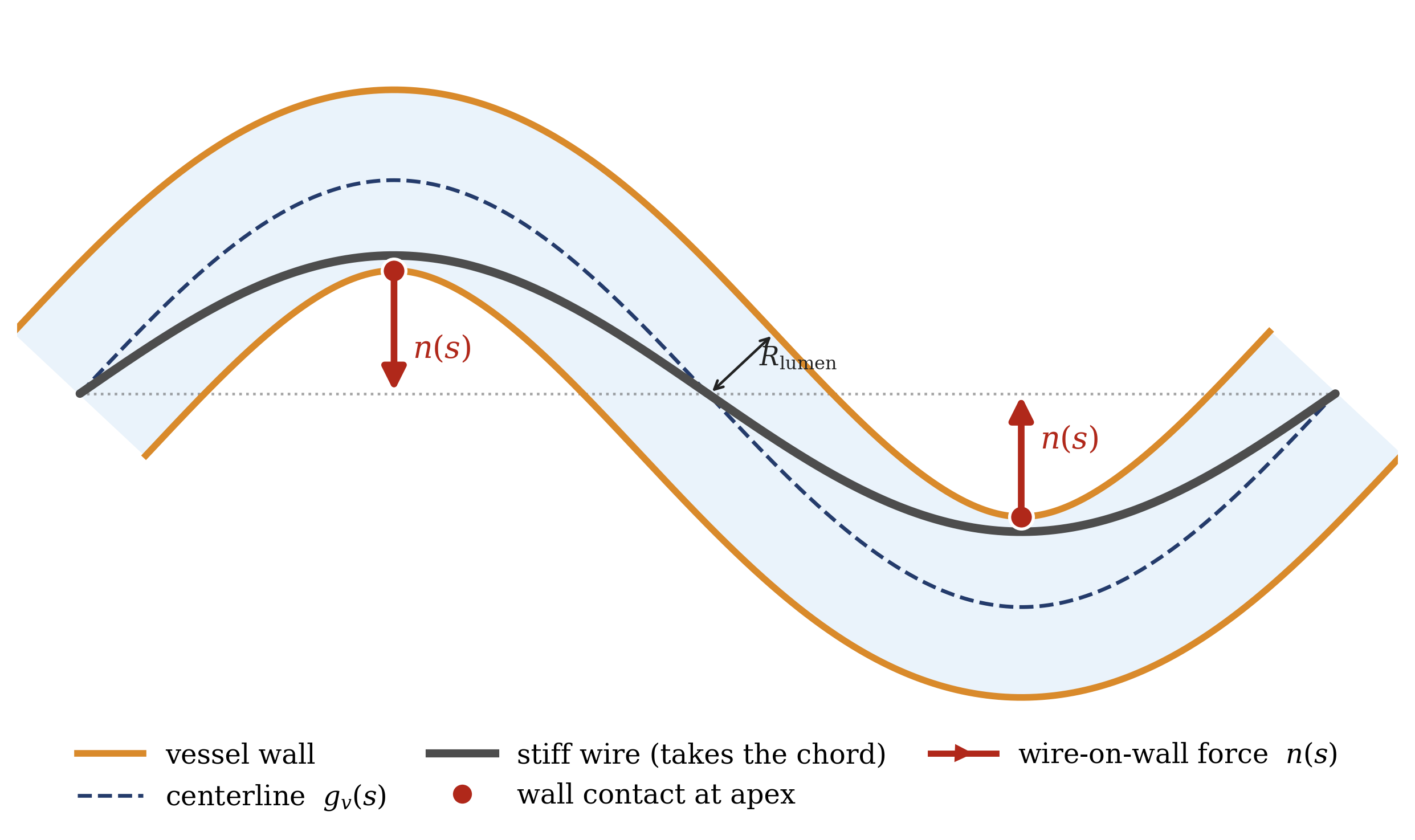}
\caption{\textbf{The unilateral lumen contact constraint, or bowstring effect.} The wire's offset from the centerline cannot exceed $\Rlum(s)-r_w$, shown here with the resulting chord-taking path and the apex contacts where the constraint binds.}
\label{fig:bowstring}
\end{figure}

The framework therefore tracks two coupled curves with a unilateral coupling. The wire's path is
constrained by the vessel's lumen surface, and the vessel's centerline is reciprocally deformed by
the wire's contact reactions. This asymmetry has direct consequences. In a tortuous iliac the wire
reaches a taut configuration before the centerline does, contact forces concentrate at the apexes
of curvature, and the \State{1}~$\rightarrow$~\State{2} mapping is not a smooth uniform
straightening but a piecewise deformation set by where the wire bowstrings and how much give exists at each contact apex.

\subsection{The same machinery bounds the vessel within the patient}
\label{sec:envelope}

The lumen inequality is one instance of a general pattern the representation makes available. Nothing developed so far prevents the vessel centerline from being displaced outside the patient. The anchoring and
extravascular terms of \S\ref{sec:Janchor} and \S\ref{sec:Jfat} resist such displacement, but they
are quadratic penalties, so they are soft priors in the sense of \S\ref{sec:hardsoft} and they
bound the displacement only through the magnitude of the loads.

Because the state is a curve of frames carrying explicit
positions $p_v(s)\in\R^3$, any anatomic region that can be given a signed distance function is
directly imposable as a second unilateral constraint of exactly the form already in use. Let
$\Omega\subset\R^3$ be an anatomic envelope segmented from the same preoperative CT that supplies
the centerline, for instance the retroperitoneal corridor or simply the soft-tissue mask, and let
$d_\Omega$ be its signed distance, negative inside. Then
\begin{equation}
d_\Omega\!\left(p_v(s)\right) \;\le\; 0 \qquad \forall s
\label{eq:envelope}
\end{equation}
enters the KKT system of \S\ref{sec:kkt} alongside Equation~\eqref{eq:closestpoint}, with its own
multipliers, its own active set, and the same complementarity structure. Anatomic containment then
becomes a \emph{hard} prior, enforced identically to lumen contact rather than penalized, and the
mobility and anchoring field is left to do what it is actually for, which is to distribute
displacement realistically within the envelope rather than to keep the vessel inside the body.

We do not implement Equation~\eqref{eq:envelope} here, since the synthetic geometries of
\S\ref{sec:numerical} have no surrounding anatomy to bound. We state it because the cost of adding
it is a property of the representation rather than of the implementation. A formulation carrying
explicit frame positions admits a region constraint immediately; one that does not carry positions,
or that represents the vessel only through a discretized wall, has no comparably direct place to
put it.

\section{Energy Functional and Anatomically Modulated Stiffness}
\label{sec:energy}

The coupled equilibrium of vessel and wire is the configuration minimizing a total elastic energy
\begin{equation}
J_{\mathrm{total}} \;=\; J_{\mathrm{wire}} + J_{\mathrm{vessel}} + J_{\mathrm{fat}}
+ J_{\mathrm{anchor}} + J_{\mathrm{axial}}
\label{eq:Jtotal}
\end{equation}
subject to the unilateral lumen inequality $h(s)\le 0$ of \S\ref{sec:contact}. This is a
constrained variational principle. The equilibrium is a configuration of stationary total stored
elastic energy that the wire-vessel system can reach without the wire penetrating the vessel wall.
The problem is manifestly non-convex, a numerical solver
returns a \emph{local} minimizer, and \S\ref{sec:numerical} documents specific spurious minima
that the geometry admits. This may not be a global minimum, and we treat determinism as
a property of the solver-plus-initialization rather than of the variational problem.

The Euler-Lagrange equations of $J_{\mathrm{total}}$ are the static Cosserat equilibrium equations on $\SE(3)$ \citep{bergou2008der,tang2012cosserat}, and after a Legendre transform they form a two-point boundary value problem in $s$. They are not a
Hamiltonian flow in any useful sense, because the unilateral constraint makes the system
non-smooth, a measure-differential inclusion rather than a smooth flow, and because
$J_{\mathrm{anchor}}$ and $J_{\mathrm{fat}}$ depend explicitly on $p_v(s)$ so the reduced system is
non-autonomous. We therefore solve the constrained minimization directly. Lie-group integrators of
Munthe-Kaas type \citep{munthekaas1998rk} remain relevant to a shooting or continuation inner loop
should stiffness require one, but they are initial-value methods and are not the right tool for
the boundary value problem as posed.

Of the five components,
$J_{\mathrm{wire}}$ and $J_{\mathrm{axial}}$ depend only on the wire path $p_w$ and its strain
$\xi_w$, while $J_{\mathrm{vessel}}$, $J_{\mathrm{fat}}$ and $J_{\mathrm{anchor}}$ depend only on
the vessel path $p_v$ and its strain $\xi_v$. \textbf{The wire-vessel system is coupled exclusively
through the lumen inequality.} Without that constraint the wire would minimize its own energy
independently and the vessel would minimize the rest independently, and there would be no
deformation at all. The constraint is what makes the problem coupled, and the contact-force
distribution at active constraints, formally the Lagrange multipliers of the KKT conditions, is
what drives the \State{1}~$\rightarrow$~\State{2} deformation. \S\ref{sec:kkt} develops this.

\subsection{Wire bending energy}
\label{sec:Jwire}

Bending a wire stores elastic energy in it, and that stored energy is what pushes it to spring back
when released. A stiff wire stores more energy per unit of bending than a soft one.
$J_{\mathrm{wire}}$ is the total stored energy as the wire deviates from its preferred, typically
straight, shape. A rod deforms in several independent ways simultaneously, bending in one
plane, bending in the perpendicular plane, twisting about its own axis, stretching axially, and
shearing. Each mode has its own stiffness coefficient, and packaging all six into a stiffness
matrix $K_w$ generalizes the scalar quadratic to a matrix quadratic $\tfrac12\,\xi^\top K_w \xi$
with $\xi\in\R^6$ the coordinates of the body-fixed strain $\hat\xi\in\se(3)$ of
\S\ref{sec:kinematics}. The energy depends on deviation from the unstressed configuration. The
rotational and translational parts of $\xi$ have different rest values. Writing
$\xi=(\omega^\top,\nu^\top)^\top$ as in \S\ref{sec:kinematics}, an unloaded rod has zero curvature
and zero twist but \emph{unit stretch}, so its rest strain is
$\xi_{w,\mathrm{intr}}=(\omega_{\mathrm{intr}}^\top,\,e_1^\top)^\top$, with
$\omega_{\mathrm{intr}}=0$ for a straight stainless-steel Lunderquist and
$\omega_{\mathrm{intr}}\ne 0$ for a pre-shaped catheter manufactured with a built-in curve. Taking
$\xi_{w,\mathrm{intr}}=0$ outright would charge an undeformed straight wire the axial energy
$\tfrac12 EA\,L_w$, which for the parameters of \S\ref{sec:params} exceeds the entire equilibrium
energy budget of \S\ref{sec:numerical} by five orders of magnitude. Separating the two blocks and
integrating over the wire length,
\begin{equation}
J_{\mathrm{wire}} \;=\; \frac12\int_0^{L_w}
\Big[\left(\omega_w-\omega_{\mathrm{intr}}\right)^{\!\top} K_\omega
\left(\omega_w-\omega_{\mathrm{intr}}\right)
\;+\; \left(\nu_w-e_1\right)^{\!\top} K_\nu \left(\nu_w-e_1\right)\Big] ds ,
\label{eq:Jwire}
\end{equation}
where $K_\omega$ and $K_\nu$ are the angular and translational blocks of $K_w$. Under the
inextensible, unshearable kinematics of \S\ref{sec:kinematics}, where $\nu\equiv e_1$, the second
bracket vanishes identically and the expression reduces to
$\tfrac12\int \omega_w^\top K_\omega\,\omega_w\,ds$, so the wire stores energy whenever its
\emph{curvature} is nonzero.

The dominant entry of $K_w$ is the bending modulus $EI$, with $E$ Young's modulus and $I$ the
second moment of area, $I=\pi r_w^4/4$ for a circular cross-section. The product relates bending
moment to curvature, $M = EI\,\kappa$. With the tangent fixed as body axis $1$, consistent with
\S\ref{sec:kinematics}, the diagonal form is
\begin{equation}
K_w \;=\; \operatorname{diag}\!\left(GI_p,\; EI,\; EI,\; EA,\; GA,\; GA\right),
\end{equation}
where $GI_p$ is the torsional rigidity with $G$ the shear modulus and $I_p = \pi r_w^4/2$ the polar second moment, and $EA, GA$ are stretch and shear stiffnesses. Under the Bishop framing of \S\ref{sec:kinematics} the material twist $\omega_1$ vanishes identically, so the $GI_p$ entry is inert and the angular block reduces to a single bending coefficient. The translational entries $EA$, $GA$ and $GA$ are likewise inert, since $\nu\equiv e_1$ removes the stretch and both shear degrees of freedom, so the operative stiffness in the models of \S\ref{sec:numerical} is the single scalar $EI$. That reduction is a consequence of the framing and kinematic choices, not an independent
physical simplification, and a torqueable guidewire in a real procedure does carry twist that this
framing does not represent.

\paragraph{On the magnitude of $EI$.} Effective bending stiffnesses quoted for clinical guidewires
are far below the nominal solid-rod value. For a Lunderquist $0.035''$ wire with
$r_w=0.4445$~mm and $E\approx 2\times10^5$~MPa, the nominal value is
$E\pi r_w^4/4 \approx 6.1\times10^3$~N$\cdot$mm$^2$, whereas effective stiffnesses in the range of
tens of N$\cdot$mm$^2$ are commonly assumed for such wires on the basis of core taper and
construction. The discrepancy is roughly two orders of magnitude and we have not found a
measurement with a stated test method that resolves it. Rather than pick a value and hope, we treat
$EI_w$ as an uncertain parameter and report in \S\ref{sec:numerical} a sweep across the full range
from $65$ to $6132$~N$\cdot$mm$^2$.

When the wire is loaded into a tortuous vessel, $\xi_w(s)$ becomes nonzero, $J_{\mathrm{wire}}$
becomes positive, and minimization of $J_{\mathrm{total}}$ trades reducing $J_{\mathrm{wire}}$,
since the wire wants to straighten, against reducing $J_{\mathrm{vessel}} + J_{\mathrm{anchor}} +
J_{\mathrm{fat}}$, since the vessel wants to stay in \State{1}. The equilibrium is the compromise,
and the ratio of $K_w$ to the vessel's restoring stiffnesses determines which side wins.

\subsection{Vessel strain energy and tortuosity modulation}
\label{sec:Jvessel}

The vessel resists deformation away from \State{1}, but its resistance is heterogeneous. The iliac
arteries in particular often carry substantial baseline tortuosity, and calcified segments are
stiff from the outset. Writing $\kappa^{\mathrm{pre}}(s)$ and $\tau^{\mathrm{pre}}(s)$ for the curvature and torsion of the preoperative centerline, a length-normalized, dimensionless tortuosity measure over the iliacs is
\begin{equation}
T_{\mathrm{tort}} \;=\; L_{\mathrm{iliac}} \int_{\mathrm{iliacs}}
\left( \kappa^{\mathrm{pre}}(s)^2 + \tau^{\mathrm{pre}}(s)^2 \right) ds .
\label{eq:tortuosity}
\end{equation}
The leading factor of segment length is deliberate. The bare integral has units of inverse length, and multiplying by $L_{\mathrm{iliac}}$ renders the index dimensionless and invariant under uniform rescaling of the geometry, so a geometrically similar but larger iliac scores the same. The clinical literature uses ``tortuosity index'' for the dimensionless arc-to-chord ratio, a different quantity;
Equation~\eqref{eq:tortuosity} is a normalized bending-energy measure and we use it only as a covariate.

The vessel strain energy is then
\begin{equation}
J_{\mathrm{vessel}} \;=\; \frac12 \int_0^{L_v}
\left(\xi_v - \xi_{v,\State{1}}\right)^{\!\top} C_v(s)
\left(\xi_v - \xi_{v,\State{1}}\right) ds ,
\label{eq:Jvessel}
\end{equation}
with $C_v(s)$ a stiffness tensor decreasing in local tortuosity and increasing in local
calcification score derived from CT. We make $C_v$ a function of \emph{local} quantities
deliberately. An earlier formulation made $C_v$ a function of the global scalar $T_{\mathrm{tort}}$, which has
the unacceptable consequence that iliac tortuosity would modulate aortic arch stiffness, and that
a focal kink and a diffuse curve with the same $T_{\mathrm{tort}}$ would produce identical stiffness everywhere. $T_{\mathrm{tort}}$ is retained as a per-case reporting covariate only.

A tortuous iliac is often described as having redundant length that gives before meaningful tension develops, after which a
much larger force is needed. That behavior is \emph{strain stiffening}, and
Equation~\eqref{eq:Jvessel} is a fixed quadratic form, hence a linear material. Making the coefficient depend on a patient-level covariate does not produce nonlinearity in strain. As
written, therefore, the model represents the honest and weaker statement that tortuous iliacs are
on average more compliant over the operative range. Representing genuine slack-then-taut behavior
requires a superquadratic energy or a bilinear $C_v$ with a knee at a slack strain $\xi_{\mathrm{slack}}(T_{\mathrm{tort}})$, which is a
straightforward extension we have not needed for the present toys.

\subsection{Extravascular resistance}
\label{sec:Jfat}

The aorta cannot translate without displacing surrounding tissue. The retroperitoneum, the
paraspinal muscle, and the spine all push back. $J_{\mathrm{fat}}$ is the elastic energy stored in
that surrounding tissue when the vessel is displaced from its preoperative position, and it is the
most patient-specific term in the model.

\paragraph{Elastic foundation.} The classical way to model a structure embedded in tissue is a
Winkler foundation \citep{hetenyi1946foundation}, treating the tissue as a continuum of independent
springs anchored at the \State{1} vessel position. To keep units unambiguous we write all
foundation terms as nodal sums rather than integrals, since the stiffnesses we quote are per-node
spring constants in N/mm,
\begin{equation}
J_{\mathrm{fat}} \;=\; \frac12 \sum_i k_{\mathrm{HU}}\!\left(\rho_{\mathrm{HU}}(x_i)\right)
\left\lVert p_v(s_i) - p_{v,\State{1}}(s_i) \right\rVert^2 .
\label{eq:Jfat}
\end{equation}
The corresponding continuum modulus in N/mm$^2$ is the nodal constant divided by the node spacing.
Mixing the two conventions is a common source of dimensional error.

\paragraph{Hounsfield density determines stiffness.} Different tissues push back differently, with
fat soft, muscle stiffer, and bone much stiffer, and each has a characteristic Hounsfield value. We therefore set $k_{\mathrm{HU}}$ from the local density through a monotone ramp with a baseline $k_0$, a reference density, and a slope $\alpha$ calibrated from a training cohort. We flag this as an author
simplification, since the density-to-modulus relationship in soft tissue is not linear in general
and a calibrated power law would be the natural refinement.

\paragraph{Where to sample, and the circularity that creates.} The centerline is the wrong place to
read density, since it sits inside the contrast-filled lumen. The sample belongs just past the
vessel wall, one to two millimeters out, in the direction the vessel is displacing. That direction
depends on the equilibrium being computed, and it is undefined at the initialization point where
the displacement is zero. We therefore hold the stiffness field \emph{fixed within each solve} and
update it between solves, so each subproblem is well posed. Three resolutions are available: a
fixed anatomic prior giving a per-segment direction from EVAR-typical anatomy, a bootstrap
iteration that recomputes the direction and re-solves to a self-consistent fixed point, and an
anisotropic stiffness tensor $K_{\mathrm{ext}}(s)$ built by sampling in many directions and
entering the energy as $\Delta p^\top K_{\mathrm{ext}}\Delta p$. The sampling geometry and the calibration of that ramp belong with the clinical imaging pipeline, which is out of scope here (\S\ref{sec:limitations}).

Only under the anisotropic option is the \emph{direction} of deformation genuinely variational.
Under the bootstrap it is a self-consistent fixed point of the sample-then-solve iteration, and
under the fixed prior it is prescribed. The clinical observation that the vessel shifts anteriorly into
retroperitoneal fat under stiff-wire forcing is therefore an \emph{output} of the anisotropic formulation and
an \emph{input} to the other two.

\subsection{Anatomic anchoring}
\label{sec:Janchor}

Local calcification and Hounsfield-derived periaortic stiffness do not capture the full structural
reality. Surgical experience is that aortic mobility is anatomically structured along the vessel's
length in a way largely shared across patients. The distal external iliacs are highly mobile. The
iliac bifurcation is moderately mobile, restrained by retroperitoneal connective tissue. The
infrarenal aorta is moderately tethered by the inferior mesenteric and lumbar arteries. The
visceral segment is anchored by the renal, superior mesenteric and celiac trunks acting as rigid
spokes. The descending thoracic aorta is tethered to the spine by the intercostals. The distal arch
and great-vessel takeoffs are essentially fixed, and the proximal arch and aortic root are anchored
to the heart. This is a property of anatomic position rather than of local image intensity, and it
is the \emph{mobility and anchoring prior} of \S\ref{sec:taxonomy}. We introduce an anatomic
anchoring stiffness $k_{\mathrm{anat}}(s)$, piecewise constant by anatomic segment, entering as
\begin{equation}
J_{\mathrm{anchor}} \;=\; \frac12 \sum_i k_{\mathrm{anat}}(s_i)
\left\lVert p_v(s_i) - p_{v,\State{1}}(s_i)\right\rVert^2 ,
\label{eq:Janchor}
\end{equation}
so that frames in heavily anchored regions are pinned near their \State{1} positions while frames
in mobile regions are free to shift under the wire's contact forces.

With regards to Equations~\eqref{eq:Jfat} and \eqref{eq:Janchor}, they are the same functional
form, and only the sum $k_{\mathrm{HU}} + k_{\mathrm{anat}}$ enters the equilibrium. They are
therefore not separately identifiable from deformation data alone, and any calibration must fix one
of them externally rather than fitting both. We keep them notationally distinct because they carry
different meanings, with $k_{\mathrm{HU}}$ capturing patient-specific variation around a population
baseline and $k_{\mathrm{anat}}$ capturing population-level structural reality, but this is an
interpretive convenience and not a claim of identifiability. Second, we have not attempted to
calibrate the anchoring profile here, and doing so against published vessel-mobility measurements
is necessary future work.

A useful subtlety emerged from the two-dimensional toy of \S\ref{sec:numerical}.
$k_{\mathrm{anat}}(s)$ should be \emph{strictly positive everywhere} along the vessel, even where
surgical intuition calls a segment free. Setting $k_{\mathrm{anat}}=0$ admits a mathematically
valid but physically degenerate equilibrium in which the vessel slides along its own arc-length
parameterization, preserving its turning-angle pattern so that $J_{\mathrm{vessel}}\approx 0$,
while drifting to a chord-aligned position with large nominal displacement and no real shape
change. Real iliacs have at minimum loose retroperitoneal tethering, and a small uniform
$k_{\mathrm{anat}}$ represents this without overconstraining the deformation. The lesson scales.
In a production model $k_{\mathrm{anat}}(s)$ takes a piecewise anatomically segmented profile but
is never zero anywhere.

\subsection{Axial pre-tension and the wire's axial degree of freedom}
\label{sec:Jaxial}

The four terms above, with the lumen inequality, do not by themselves produce clinical-magnitude
vessel deformation. The wire's bending energy at realistic $EI_w$ generates only sub-newton contact
forces against the wall, deforming the vessel by a fraction of a millimeter, far less than is seen
on intraoperative angiography. The missing mechanism is axial loading of the wire.

In real EVAR the operator pushes and pulls the wire as it is advanced from the femoral access
toward the aortic arch, so the wire carries non-trivial axial tension $T_{\mathrm{axial}}$ along its
length. Cosserat-rod-in-tube mechanics gives the consequence. An axially tensioned wire traversing
curvature $\kappa$ exerts a transverse line load on the lumen wall,
\begin{equation}
\lambda(s) \;=\; T_{\mathrm{axial}}\,\kappa(s),
\label{eq:lineload}
\end{equation}
which is the transverse equilibrium of a tensioned string. It is not the capstan relation; the
capstan equation $T_2 = T_1 e^{\mu_f \beta}$ is a friction law, and we return to that distinction
below. We model the work the axial
tension does as the wire deviates from a straight chord between its endpoints,
\begin{equation}
J_{\mathrm{axial}} \;=\; T_{\mathrm{axial}}
\left( L^{\mathrm{arc}}_w(p_w) - L^{\mathrm{chord}}_w \right),
\label{eq:Jaxial}
\end{equation}
with $L^{\mathrm{arc}}_w$ the wire's instantaneous arc length and $L^{\mathrm{chord}}_w =
\lVert p_w(L_w)-p_w(0)\rVert$ the straight-line distance between its endpoints. Minimizing
$J_{\mathrm{axial}}$ pulls the wire toward the chord and the lumen constraint resists. Both wire
endpoints are held (\S\ref{sec:phase1}), so $L^{\mathrm{chord}}_w$ is a constant and the
subtraction is a fixed-endpoint offset that makes the term nonnegative rather than part of the
physics. Were an endpoint free, the $-T_{\mathrm{axial}}L^{\mathrm{chord}}_w$ term would contribute
a spurious end force of magnitude $T_{\mathrm{axial}}$ along the chord, and the term should then be
written as $T_{\mathrm{axial}}L^{\mathrm{arc}}_w$ up to a constant.

\paragraph{Contact is assumed frictionless.} No tangential force appears in the contact set of
\S\ref{sec:contact}, and $T_{\mathrm{axial}}$ is accordingly constant along $s$. Physically the
load enters at $s=0$, the common femoral access, where the sheath holds the wire. Coulomb friction
would make the tension decay cranially as the wire wraps each curvature apex, with $\lvert
dT/ds\rvert \le \mu_f\lambda(s)$, the capstan relation proper, and $T$ approaching zero near the
parked tip. Neglecting friction is a modeling assumption that over-estimates the tension delivered
to the distal segment, and it is the natural place at which a traction boundary condition would
enter once friction is modeled. We flag it here rather than model it.

\paragraph{The wire slides.} This term only does work if the wire's in-vessel arc length can
change. If the wire were treated as
inextensible \emph{and} its endpoints held, then both $L^{\mathrm{arc}}_w$ and
$L^{\mathrm{chord}}_w$ would be fixed, $J_{\mathrm{axial}}$ would be constant, and it would exert
no force whatsoever. An earlier version of this work resolved that tension by adding a per-segment
Hookean inextensibility spring soft enough to let the wire shorten. That is not defensible. The
true axial stiffness of a steel guidewire is $EA/\ell_0 \sim 10^5$~N/mm per millimeter segment,
several orders of magnitude stiffer than any spring that would permit the observed shortening, so
such a spring is a numerical device masquerading as physics, and it dominated the energy budget.

The correct physical picture is that the wire is not a closed system. It runs out through the
sheath to the operator's hand, so the intraluminal arc length is genuinely free while the wire
itself remains inextensible, and $T_{\mathrm{axial}}$ is a dead load. We adopt that picture.
The wire's total arc length is an unconstrained degree of freedom, no inextensibility energy
appears in Equation~\eqref{eq:Jtotal}, and the node-bunching that motivated the spring is handled
where it belongs, as a \emph{parameterization} constraint requiring uniform node spacing, which
contributes nothing to the energy. \S\ref{sec:numerical} quantifies how much this changes the
answer, and the change is substantial.

$T_{\mathrm{axial}}$ is the fifth physically meaningful parameter family of the forward model. It
is set by the surgical push and pull, it is the one parameter not derivable from preoperative
imaging, and we treat a range of order $2$ to $15$~N as a working assumption pending calibration
against intraprocedural force-sensor data. We flag it explicitly as an assumption because
\S\ref{sec:numerical} finds it to be the dominant energetic driver, and a dominant parameter
resting on an assumption is exactly the thing a reader should be told about.

\subsection{Numerical enforcement of the lumen contact constraint}
\label{sec:kkt}

The equilibrium problem is the constrained minimization
\begin{equation}
\min_{p_w,\,p_v}\; J_{\mathrm{wire}} + J_{\mathrm{vessel}} + J_{\mathrm{fat}}
+ J_{\mathrm{anchor}} + J_{\mathrm{axial}}
\qquad \text{subject to}\qquad h(s)\le 0\ \ \forall s .
\label{eq:minproblem}
\end{equation}

\paragraph{KKT formulation.} The Karush-Kuhn-Tucker conditions introduce a multiplier field $\mu(s)\ge 0$ at each arc length and require complementarity $\mu(s)h(s)=0$ at the optimum, so $\mu(s)>0$ only where the wire is in active contact. This multiplier is the contact force magnitude of Equation~\eqref{eq:complementarity}, the physical contact force conjugate to the gap. We reserve the symbol $\lambda$ for the line load of Equation~\eqref{eq:lineload}, which carries different units. The contact force is therefore the variational dual of the constraint rather than a separate object to be modeled, and it falls out of the solver. Discrete multipliers for nodal constraints carry units of force in N, conjugate to a gap in mm, whereas the line load of
Equation~\eqref{eq:lineload} is in N/mm, so converting between them requires dividing by the node
spacing.

\paragraph{Equilibrium interpretation.} The stationarity condition for $p_v$ balances the vessel's
elastic and tissue restoring force against the contact force from the wire,
\begin{equation}
\frac{\partial\left(J_{\mathrm{vessel}}+J_{\mathrm{fat}}+J_{\mathrm{anchor}}\right)}{\partial p_v(s_i)} \;=\; \mu_i\,\hat r(s_i),
\end{equation}
written nodally, consistent with the convention of \S\ref{sec:Jfat}, with $\mu_i$ the nodal contact force in newtons and $\hat r(s_i)$ the unit vector from vessel to wire. Where the inequality is slack the right side vanishes and the vessel is in free elastic equilibrium under its neighbors at that station; it returns to \State{1} only where the constraint is inactive over a neighborhood large compared with the elastic decay length, since $J_{\mathrm{vessel}}$ couples each node to its neighbors. Where it is tight the wire pushes outward and the vessel
deforms until its restoring force balances the push. By Newton's third law the wire feels $-\mu_i\hat r(s_i)$, which bends it away from the chord. This is the entire mechanism of the
\State{1}~$\rightarrow$~\State{2} deformation.

\paragraph{Which reaction to report.} The nodal multipliers are in fact determined, since each lumen inequality involves its own wire node and the active constraint gradients are therefore linearly independent. They are, however, \emph{extensive}: $\mu_i \approx \lambda(s_i)\,\Delta s_i$, so an individual multiplier depends on the discretization, and it is noisy at an active-set boundary where the local Voronoi length collapses. We therefore report the smooth line load
$\lambda(s)=T_{\mathrm{axial}}\kappa(s)$ of Equation~\eqref{eq:lineload}, restricted to the active
set, as the physical contact reaction. Two caveats attach to that choice. First, restricting to
the active set is essential, because $T\kappa$ is nonzero wherever the wire is curved, including
stretches where it is not touching the wall and the true reaction is zero by definition. Second,
$T\kappa$ is the transverse load on a tensioned string, and the full elastica balance carries an
additional bending correction of order $EI(\kappa'' + \tfrac12\kappa^3)$, which is small at the low
end of the $EI_w$ range and not negligible at the high end.

\paragraph{Penalty and augmented-Lagrangian alternatives.} When the constraint count is too large for active-set methods, the inequality can be replaced by a soft quadratic penalty on $\max(0,h)$, which recovers the hard constraint in the limit of large penalty weight but generates force only \emph{after} interpenetration, so the equilibrium always carries a residual gap error. The augmented Lagrangian removes that bias by carrying an explicit multiplier estimate and converges to the true KKT multiplier without an impractically large weight. We use a KKT active-set method for the toys of \S\ref{sec:numerical} and keep the augmented-Lagrangian path open for production scale.

\paragraph{The soft-prior counterpart.} The same inequality can instead be folded into the outer
AINN training loss as a soft prior,
\begin{equation}
\Lcal_{\mathrm{lumen}}(\theta) \;=\; \int_0^{L_w}
\max\!\left(0,\; d_c(s;\theta) - \left(\Rlum(\sigma^*(s))-r_w\right)\right)^2 ds ,
\end{equation}
which is the loss-function realization of the contact and exclusion prior category of
\S\ref{sec:taxonomy}. In this worked example we do \emph{not} use $\Lcal_{\mathrm{lumen}}$, and
enforce contact as a hard architectural prior inside the simulator's KKT solve. It is documented
here as the natural soft alternative for AINN applications where the constraint is approximate or
stochastic and a hard constraint would be too restrictive.

\subsection{Parameter summary}
\label{sec:params}

The forward model has five physically meaningful parameter families, together with one reporting covariate, summarized in Table~\ref{tab:params}. Four of the five are derivable from preoperative imaging together with catheter selection and anatomic segmentation, and require no additional intraoperative measurement at inference time. The fifth is the axial pre-tension of \S\ref{sec:Jaxial}, which would be refined intraoperatively where force-sensor data are available.

\begin{table}[t]
\centering
\small
\caption{Parameter families of the forward model. $T_{\mathrm{tort}}$ is listed separately because \S\ref{sec:Jvessel} demotes it from a model parameter to a per-case reporting covariate.}
\label{tab:params}
\begin{tabular}{@{}>{\raggedright\arraybackslash}p{0.20\textwidth}>{\raggedright\arraybackslash}p{0.38\textwidth}>{\raggedright\arraybackslash}p{0.34\textwidth}@{}}
\toprule
\textbf{Parameter} & \textbf{Mathematical role} & \textbf{Clinical proxy} \\
\midrule
$K_w$ (or $K_{\mathrm{eff}}$) & Bending stiffness of wire, or of the wire-and-device assembly, as a quadratic form on $\se(3)$ & Lunderquist versus Amplatz versus Glidewire; device deployment state \\[3pt]
$C_v(s)$ & Local anisotropic vessel stiffness tensor & Calcification score, wall quality \\[3pt]
$k_0,\alpha$ & Coefficients of the Hounsfield-to-stiffness ramp feeding
$k_{\mathrm{HU}}$ & BMI, periaortic fat density \\[3pt]
$k_{\mathrm{anat}}(s)$ & Anatomic-position anchoring stiffness & Arch, visceral, retroperitoneal and iliac segments \\[3pt]
$T_{\mathrm{axial}}$ & Dead load driving the wire toward its chord & Surgical push and pull; intraprocedural force sensing \\
\midrule
\multicolumn{3}{@{}l@{}}{\emph{Reporting covariate, not a model parameter}} \\[2pt]
$T_{\mathrm{tort}}$ & Normalized bending-energy measure of iliac slack & Iliac tortuosity \\
\bottomrule
\end{tabular}
\end{table}

\section{The Optimal-Transport Loss}
\label{sec:wasserstein}

Minimizing $J_{\mathrm{total}}$ produces a \emph{forward} equilibrium prediction of \State{2}. Two
questions remain. How do we measure how well that prediction matches intraoperative reality, and
how, given a population of patients with paired data, do we \emph{learn} a refined predictive
operator that does better than the bare mechanical simulation?

A Euclidean mean-squared error between predicted and observed point clouds is structurally
inappropriate for vascular anatomy. It is sensitive to nuisance parameterization, since which
centerline point corresponds to which is arbitrary. It ignores topological identity, so it cannot
tell a correctly branching vessel from an incorrect one. And it assigns \emph{low} error to some
anatomically impossible configurations, such as a prediction that doubles back on itself while
remaining pointwise close to the target.

\subsection{Wasserstein-2 between measures of frames}

We use instead the Wasserstein-2 distance between probability measures defined on the manifold
$\SE(3)$. The clinical intuition is the earth-mover's distance. Imagine the \State{1} vessel tree
as a pile of dirt distributed along the centerlines, and the \State{2} tree as a different pile at
the deformed positions. The Wasserstein-2 distance is the minimum total work needed to shovel the
first pile into the shape of the second, where the cost of moving each grain is proportional to the
squared distance moved. Two trees that look mostly the same but differ near the iliac bifurcation
cost less to transport than two trees that differ everywhere. $W_2$ matches
distributions of mass rather than fixed index pairs, so it is far less sensitive to nuisance
parameterization than pointwise error.

Modeling the \State{1} and \State{2} trees as probability measures $\mu_1,\mu_2$ of frames along
the vessel tree, each frame weighted for instance by local lumen cross-sectional area,
\begin{equation}
W_2^2(\mu_1,\mu_2) \;=\; \inf_{\gamma\in\Gamma(\mu_1,\mu_2)}
\int_{\SE(3)\times\SE(3)} d^2(x,y)\; d\gamma(x,y),
\label{eq:w2}
\end{equation}
where $\Gamma(\mu_1,\mu_2)$ is the set of couplings with the prescribed marginals and $d$ is a
distance on $\SE(3)$. The coupling $\gamma$ is a transport plan specifying, for every grain of dirt
in $\mu_1$, how much goes to each location in $\mu_2$. The marginal constraint means every grain
must be moved and the destination pile must come out right, the integral measures the total cost of
the plan, and the infimum picks the cheapest plan.

We should be careful here to state what $W_2$ does and does not deliver. It mitigates
parameterization sensitivity. It does \emph{not} by itself enforce branch identity, for which a
topology-aware objective using persistence diagrams or a fused Gromov-Wasserstein variant would be
required \citep{hu2019topology,byrne2023topological}. Nor does it exclude anatomically impossible
configurations on its own. In this framework that role is filled by the $\SE(3)$ representation and
the contact constraint, which are hard priors, not by the loss.

\paragraph{A caution about mass and foreshortening.} $W_2$ requires equal total mass, so the
measures must be normalized. But foreshortening, meaning a change in total arc length, is one of
the clinical phenomena we most want to capture, and normalizing removes it from the loss entirely.
One solution is unbalanced optimal transport with an explicit mass-creation
penalty \citep{chizat2018unbalanced}. The other, simpler and adequate here, is to normalize and add
a scalar length term $\bigl|\hat L_v - L_v^{\mathrm{obs}}\bigr|$ so that foreshortening is
supervised directly. We adopt the latter.

\subsection{Ground cost}
\label{sec:groundcost}

The natural ground cost on $\SE(3)$ is not Euclidean. A frame cannot translate sideways
instantaneously any more than a car can slide laterally, and the sub-Riemannian, or
Carnot-Carathéodory, distance encodes exactly this non-holonomic structure as the length of the
shortest curve whose tangent everywhere lies in an admissible horizontal distribution. Taking the distribution spanned by forward translation and the three rotations, the missing lateral directions are recovered by bracketing, so the resulting distance is finite and induces the manifold topology, which is what makes $W_2$ with this ground cost a genuine metric on the space of measures. Optimal transport on roto-translation groups with left-invariant and sub-Riemannian
ground metrics has been developed in detail for $\SE(2)$ by \citet{bon2025otse2}, and
sliced-Wasserstein constructions exist for other manifold classes \citep{bonet2025chmanifolds}, though $\SE(3)$ falls outside the setting those assume.

The supervision available to us is, however, a single projected view, and the projection of
\S\ref{sec:projection} extracts only the position component of each frame. The loss that trains
the network is therefore computed between measures on $\R^2$ with a Euclidean ground cost. The
sub-Riemannian machinery is not exercised by the training gradient.

Second, and following from this, the rotational half of the residual correction receives
essentially no gradient from the data term. Writing the residual as $\hat\eta_\theta = (\omega_\eta,\nu_\eta)$, the position component of the composed frame of \S\ref{sec:architecture} is $p(\hat g) = p_{\mathrm{sim}} + R_{\mathrm{sim}} V(\omega_\eta)\,\nu_\eta$, where $V(\cdot)$ is the left Jacobian of $\SO(3)$ appearing in the closed-form $\SE(3)$ exponential. Its derivative with respect to $\omega_\eta$ is proportional to $\nu_\eta$ and vanishes exactly where the translational residual does. The three rotational degrees of freedom per node are therefore only weakly identified from monoplane data, through a term of order $\lVert\nu_\eta\rVert$ whose Jacobian degenerates as the residual approaches the physics prediction, and they are otherwise constrained only by the smoothness regularizer of \S\ref{sec:architecture}, which penalizes the arc-length derivative and therefore leaves a constant rotational offset in its null space. Monoplane supervision separately leaves the depth component of each predicted position unobserved, which \S\ref{sec:depth} treats on its own. We regard this as a property of the data, not a defect of
the method, but it must be stated rather than obscured.

Third, our numerical work in \S\ref{sec:numerical} evaluates a nilpotent first-order approximation
to the sub-Riemannian distance, not the exact Carnot-Carathéodory distance. A nilpotent
approximation is a local model and need not satisfy the triangle inequality globally.

So the next goal was to try to simplify things. In the forward model as implemented, every energy term and the
contact inequality are functionals of the two position curves. Under the Bishop framing the bending
energies reduce to $\tfrac12\int EI\kappa^2\,ds$, and Equation~\eqref{eq:closestpoint} is a minimum
over Euclidean distances, so the mechanics of \S\ref{sec:numerical} could be reproduced in $\R^3$.
What the group buys is four things, none of them the loss. The incremental
twist $\Log(g_i^{-1}g_{i+1})/\Delta s_i$ estimates bending without degenerating at the
near-zero-curvature points where a finite-differenced Frenet frame does, which is the regime of a
stiff wire. Composition on the group puts the hard geometric prior of \S\ref{sec:architecture} in the learned
half of the model rather than the physics half, since it is $g_{\mathrm{sim}}\Exp(\hat\eta_\theta)$
that cannot leave the manifold, and that construction has no $\R^3$ analogue. Reading out
$d_{\text{in-frame}}$ requires a frame by definition, since it lives in the wire's own
cross-section. And every extension we identify as necessary for clinical fidelity, meaning an
anisotropic $C_v$, a torqueable wire carrying material twist, and a pre-shaped catheter with
$\omega_{\mathrm{intr}}\ne 0$, is a rotational quantity that $\R^3$ cannot carry and that the
present models switch off. The claim is that $\SE(3)$ is the right container, not that the forward
model as implemented would fail without it. The
sub-Riemannian ground cost is the mathematically natural choice and becomes operative in the
three-dimensional setting, for instance when a completion cone-beam CT provides true 3D ground
truth for a held-out subset. We flag its use as a direction rather than presenting it as the
methodological core of what is verified here.

\subsection{Computing \texorpdfstring{$W_2^2$}{W2} in practice}

The minimization in Equation~\eqref{eq:w2} is a linear program in the transport plan, the
Kantorovich primal. Discretizing the predicted and observed measures into finite point masses
$\{x_i,a_i\}_{i=1}^m$ and $\{y_j,b_j\}_{j=1}^n$,
\begin{equation}
W_2^2(\mu_1,\mu_2) \;=\; \min_{\gamma_{ij}\ge 0}\ \sum_{i,j}\gamma_{ij}C_{ij},
\qquad C_{ij}=d^2(x_i,y_j),
\end{equation}
subject to $\sum_j \gamma_{ij}=a_i$ and $\sum_i \gamma_{ij}=b_j$. This is a standard LP with $mn$
variables and $m+n$ equality constraints, of which one is redundant. For landmark measures with
$m,n\sim10$ and wire-path measures with $m,n\sim10^2$ it is solved exactly in well under a second.
Because our pipeline runs preoperatively and is not time-critical, we prefer the exact LP to the
entropic-regularized Sinkhorn iteration \citep{cuturi2013sinkhorn}, which is faster and gives
smoother gradients but introduces a regularization parameter that biases the estimate. We would
revisit that choice for real-time deployment or for dense-measure training at scale, where the
LP's cubic-ish scaling does become limiting. At $m=n\sim10^3$ the LP has $10^6$ variables and is no
longer a sub-second computation inside a training loop.

One implementation note matters for the gradient below: the LP must be solved by a simplex-type method, since the deterministic matching used below requires a vertex solution and an interior-point solver returns a valid but dense plan.

\paragraph{Differentiability.} By Danskin's theorem the LP value is differentiable in the cost
matrix wherever the optimal plan $\gamma^\star$ is unique, with $\partial W_2^2/\partial C_{ij} =
\gamma^\star_{ij}$, and uniqueness holds for generic $C$, hence almost everywhere. For a
squared-Euclidean ground cost this gives
\begin{equation}
\nabla_{x_i} W_2^2 \;=\; 2\sum_j \gamma^\star_{ij}\left(x_i-y_j\right),
\end{equation}
which reduces to $2a_i(x_i-y_{j^\star(i)})$ only when the plan is deterministic at $i$. The
descent direction is the negative of this. Degenerate configurations, where the optimum is not
unique, are exactly the symmetric ones, and uniformly arc-length-sampled polylines can produce
them, so a small jitter or a brief entropic smoothing is advisable at those points. This loss can
therefore be backpropagated through a downstream network.

\section{Single-View Projection and 2D Supervision}
\label{sec:projection}

The loss of \S\ref{sec:wasserstein} is specified between two measures that are implicitly
three-dimensional. In practice the intraoperative ground truth at most centers is a single-view 2D
fluoroscopic angiogram captured with the stiff wire and the partially or fully deployed delivery
system in place. Cone-beam CT and biplane fluoroscopy exist in some hybrid rooms but are not
routinely captured. The framework must therefore train against weaker, projected supervision while
continuing to predict in three dimensions.

\subsection{The projection operator}

We compute the loss by \emph{pushing forward} the predicted 3D measure through the C-arm projection
and comparing in 2D. Let $T_C\in\SE(3)$ denote the C-arm pose for the angiographic run, meaning
gantry primary and secondary angles together with source-to-image and source-to-object distances
and the principal point, all reconstructible from the acquisition metadata. Let
$\pi:\R^3\to\R^2$ be the perspective point-source projection of a pinhole camera,
$\pi(x,y,z) = (f x/z,\, f y/z)$ with $f$ the effective focal length. Writing $p(g)$ for the
position component of a frame, the composite operator is
\begin{equation}
\Pi \;=\; \pi \circ T_C^{-1} \circ p, \qquad \Pi:\SE(3)\to\R^2 ,
\label{eq:projop}
\end{equation}
acting on a frame by extracting its position, transforming into the C-arm coordinate frame, and
projecting through the imaging pinhole. The 2D-supervised loss is then
\begin{equation}
\Lcal_{\mathrm{2D}}(\theta) \;=\; W_2^2\!\left(\Pi_\sharp F_\theta(\State{1}),\ \mu_{\mathrm{fluoro}}\right),
\end{equation}
where $\Pi_\sharp$ is the pushforward of the predicted measure to a measure on $\R^2$ and
$\mu_{\mathrm{fluoro}}$ is the empirical measure constructed from the angiogram. As noted in
\S\ref{sec:groundcost}, Equation~\eqref{eq:projop} discards the rotation component of $g$
entirely, so the ground cost on $\R^2$ is the standard Euclidean cost and the geometric structure of
$\SE(3)$ has done its work in the forward model before the prediction is projected.

\subsection{What the angiogram supplies}

Full 2D centerline extraction from a fluoroscopic angiogram is hard, because the contrast bolus is
transient and noisy, the wire and device project through the lumen, and other anatomy projects
through the field. Fortunately a full deformed-vessel centerline is not needed. The angiogram
supplies two cleaner pieces of information.

\begin{enumerate}[leftmargin=1.8em,itemsep=3pt,topsep=3pt]
\item \textbf{The projected wire path}, a continuous 2D curve $\Pi(g_w(s))$. The stiff wire is
radio-opaque and persistent across all frames of the run, making it the most reliable object on the
image, and it is exactly the projected version of the wire path the model predicts, so the
comparison is like for like.
\item \textbf{A small set of clinically meaningful 2D landmarks.} Of these only the lowest renal
ostium and the aortic bifurcation are reliably identifiable on essentially every case; the SMA, the
celiac trunk, the iliac bifurcations, the wire tip, and the device's proximal and distal
radio-opaque markers are used when visible.
\end{enumerate}

The loss is therefore a sum of two Wasserstein terms, one continuous for the wire path and one
discrete for the landmarks,
\begin{equation}
\Lcal_{\mathrm{2D}}(\theta) \;=\; w_{\mathrm{wire}}\, W_2^2\!\left(\Pi_\sharp \hat\mu_w,\ \mu_{\mathrm{wire}}\right)
\;+\; w_{\mathrm{lm}}\, W_2^2\!\left(\Pi_\sharp \hat\mu_{\mathrm{lm}},\ \mu_{\mathrm{lm}}\right),
\label{eq:L2D}
\end{equation}
with $\hat\mu_{\mathrm{lm}}$ the predicted measure supported on the model's landmark positions and $w_{\mathrm{wire}}, w_{\mathrm{lm}}$ weights reflecting the relative reliability of the two sources. Both targets are
matched to like objects, the predicted wire's projection against the observed wire curve and
predicted landmark points against observed landmark points, so the transport never has to map a
thin predicted structure onto a thick contrast silhouette, which would be ill-posed. We
deliberately supervise on the radio-opaque wire and the discrete landmarks rather than on the
transient contrast blush for exactly this reason. Where a filled-lumen silhouette is preferred, the
predicted centerline can first be rendered as a tube of radius $\Rlum(s)$ and projected, so both
measures are two-dimensional before transport.

\subsection{Depth-axis ambiguity}
\label{sec:depth}

Projection onto a single image plane drops one dimension of information per landmark. The working
view in EVAR is not a pure anteroposterior image but whatever oblique and cranio-caudal angulation
best displays the lowest renal ostium, and that angulation is recorded, so the projection geometry
is known. Nonetheless a prediction that is correct in the working view but wrong along the depth
axis is scored as correct by $\Lcal_{\mathrm{2D}}$. This is a fundamental limitation of monoplane
fluoroscopy and is the reason 2D-supervised 3D prediction is often considered ill-posed.

The architecture of \S\ref{sec:architecture} is designed to address it. The physics simulation
produces a 3D prediction whose depth is determined by the mechanics of the wire-vessel
interaction, the 2D loss disambiguates the in-plane components, and the residual is regularized
along the unobserved direction. Concretely we penalize the depth component of the position change
induced by the residual,
\begin{equation}
\Lcal_{z} \;=\; \lambda_z \int_0^{L_v}
\left( e_z^{\top} R_C^{\top} \left[\, p(\hat g_v(s)) - p(g_{\mathrm{sim}}(s)) \,\right] \right)^{2} ds ,
\label{eq:depthpen}
\end{equation}
where $g_{\mathrm{sim}}$ denotes the physics-simulator prediction of \S\ref{sec:architecture}, $R_C$ is the rotation of $T_C$, and $e_z$ is the detector-frame depth direction. We write it this
way because the naive form, pairing a three-vector with a Lie-algebra element, is not well typed,
and because the depth axis is a detector-frame direction while the residual is a body-frame algebra
element, so a frame-free pairing does not exist.

We should be measured about what this buys. Equation~\eqref{eq:depthpen} is a quadratic penalty on
an unobserved component, which is Tikhonov regularization; what differs from generic smoothness
regularization is the \emph{prior}, since the out-of-plane component is supplied by a mechanical
model rather than by an assumption of smoothness. Whether a mechanical prior outperforms the
alternatives is an empirical question and not something we have tested. Existing monoplane
approaches in this setting resolve depth using wire-shape priors \citep{breininger2019stiffwires}
or learned statistical shape models \citep{zhou2018instantiation}, and any claim of superiority
would have to be made against those, not against a strawman.

\subsection{The wire-and-device assembly}
\label{sec:assembly}

The angiogram captures the deformed vessel with both the stiff wire and the partially deployed
delivery system in place, and both contribute mechanically. The framework treats the assembly as a
single coupled stiff inclusion with effective stiffness $K_{\mathrm{eff}}(s)$ and effective radius
$r_{\mathrm{eff}}(s)$, both varying with arc length depending on whether the inclusion at $s$ is the
bare wire or the wire-loaded sheath. The contact constraint generalizes with $r_w$ replaced by
$r_{\mathrm{eff}}(s)$ and the bending term of \S\ref{sec:Jwire} uses $K_{\mathrm{eff}}$.

This generalization has a limit worth stating. A large-bore sheath can have an outer radius
comparable to, or exceeding, the lumen radius of a small external iliac, at which point
$r_{\mathrm{eff}} \ge \Rlum$ and the feasible set of the rigid-wall problem is empty. That is not a
pathological corner case but precisely the clinically dangerous situation of iliac conduit
difficulty and rupture risk. Representing it requires a compliant wall, replacing $\Rlum(\sigma)$
by $\Rlum(\sigma)+u(\sigma)$ with a hoop energy $\tfrac12 k_{\mathrm{hoop}} u^2$, so that
oversizing produces wall strain rather than infeasibility. The present model assumes a rigid wall
and is therefore restricted to $r_{\mathrm{eff}} < \Rlum$.

\section{Architecture: a Physics Prior with an \texorpdfstring{$\SE(3)$}{SE(3)}-Equivariant Residual}
\label{sec:architecture}

A neural operator trained from scratch on a projected optimal-transport loss with a few dozen
patient pairs will overfit. The Wasserstein loss makes predictions geometrically well behaved, but
it does not make them generalize. The supervision is weak, being one projection per case, the
cohort is small, and the underlying map is nonlinear in several parameter families. The remedy is
structural. We put the physics in the architecture, not only in the loss
(Figure~\ref{fig:architecture}).

\subsection{Composition on the manifold}

The forward physics simulation of \S\ref{sec:energy} produces a deterministic prediction, and a
learned network produces a small Lie-algebra-valued correction composed multiplicatively with it,
\begin{equation}
\hat g_v(s) \;=\; \mathrm{PhysicsSim}(\State{1},\mathbf{p})(s)\;\cdot\;
\Exp\!\left(\hat\eta_\theta(s)\right),
\label{eq:composition}
\end{equation}
where $\mathbf{p}$ collects the physical parameters of \S\ref{sec:params} and
$\hat\eta_\theta(s)\in\se(3)$ is the network's residual output at frame $s$. Composition with the
matrix exponential ensures the corrected prediction stays on $\SE(3)$ by construction, since
$\SE(3)$ is closed under multiplication and $\Exp$ maps any Lie-algebra element into the group. The
correction can only rotate and translate frames in physically realizable ways, regardless of what
the network does.

\paragraph{Which side, and what transforms how.} The multiplication in
Equation~\eqref{eq:composition} is on the \emph{right}, and the choice is not arbitrary. For the
composition to be left-equivariant under a global rigid motion $q\in\SE(3)$, meaning $\hat g_v(q\cdot\State{1}) = q\cdot\hat g_v(\State{1})$, we need $q\,g_{\mathrm{sim}}\Exp(\hat\eta_\theta)$, which requires the residual to be \emph{invariant}, $\eta_\theta(q\cdot x)=\eta_\theta(x)$, rather than equivariant.
Right-multiplication is correct precisely because the residual lives in the body frame. We
therefore produce $\eta_\theta$ as an invariant function of body-frame features, and the
composition is then left-equivariant.

\paragraph{Where equivariance can break.} The composition inherits equivariance only if the
simulator is equivariant too, and this is a real caveat rather than a formality. The
fixed-direction option for the extravascular sampling of \S\ref{sec:Jfat}, which hard-codes an
anterior or anterolateral displacement direction, is expressed in the world frame and therefore
breaks simulator equivariance. Equivariance holds only if that anisotropy direction is expressed
in a patient-attached frame that transforms with the anatomy. We state this because it is the kind
of detail that quietly invalidates an equivariance claim.

\paragraph{What the composition preserves, and what it does not.} Every corrected frame remains in $\SE(3)$, so the
per-frame guarantee of \S\ref{sec:kinematics} survives, but the chain relation does not. Adjacent
frames are perturbed independently, so the reconstructed segment lengths are no longer carried by
the representation, and the lumen inequality $h(s)\le 0$, enforced hard inside the equilibrium
solve, is not re-imposed after composition. A large enough residual can therefore push the wire
through the wall. The architecture as specified discards two guarantees the forward model has, and
we flag this rather than repair it here. Two remedies are available. Applying the residual in the strain domain, correcting $\xi_i$ and reconstructing by exponentiation, restores the chain relation and with it controlled arc length. Re-projecting the composed prediction onto $\{h\le 0\}$, or
placing the residual inside the equilibrium solve rather than after it, restores feasibility. The
smoothness penalty below is a soft substitute for the first of these, not a replacement.

\paragraph{Smoothness.} The physics prediction is smooth by construction, since the Cosserat-rod
equilibrium penalizes bending, but a high-frequency residual could inject a non-physical kink into
the composed curve. We therefore regularize $\eta_\theta$ with a roughness penalty
$\lambda_{\mathrm{smooth}}\int \lVert d\eta_\theta/ds\rVert^2 ds$, equivalently a penalty on the
difference between adjacent-frame residuals or a band-limited parameterization, so the corrected
prediction inherits the continuity of the equilibrium it perturbs.

\subsection{The complete training objective}

Collecting the pieces, which are otherwise scattered across sections, the objective is
\begin{equation}
\Lcal(\theta,\mathbf{p}_{\mathrm{cal}}) \;=\;
\underbrace{w_{\mathrm{wire}} W_2^2\!\left(\Pi_\sharp \hat\mu_w, \mu_{\mathrm{wire}}\right)
+ w_{\mathrm{lm}} W_2^2\!\left(\Pi_\sharp \hat\mu_{\mathrm{lm}}, \mu_{\mathrm{lm}}\right)}_{\text{data}}
\;+\; \underbrace{\lambda_L \left| \hat L_v - L_v^{\mathrm{obs}} \right|}_{\text{foreshortening}}
\;+\; \underbrace{\Lcal_z}_{\text{depth}}
\;+\; \underbrace{\lambda_{\mathrm{smooth}}\!\int\!\left\lVert \tfrac{d\eta_\theta}{ds}\right\rVert^2\! ds}_{\text{residual smoothness}} ,
\end{equation}
minimized jointly over the network weights $\theta$ and the calibration parameters
$\mathbf{p}_{\mathrm{cal}}$.

\begin{figure}[t]
\centering
\includegraphics[width=\textwidth]{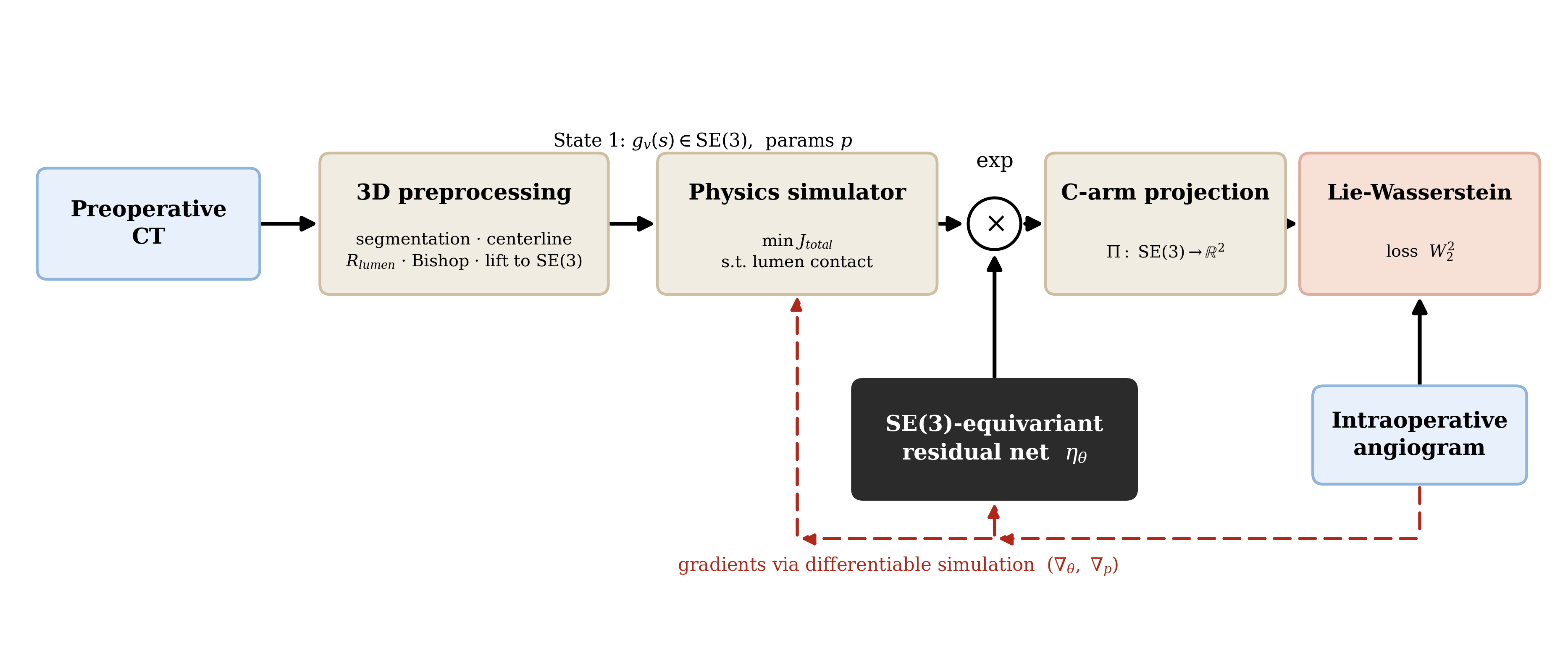}
\caption{\textbf{End-to-end AINN architecture for the worked example.} Preoperative CT feeds a 3D
preprocessing pipeline producing the \State{1} representation and the patient-specific physical
parameters. The \S\ref{sec:energy} physics simulator gives a deterministic forward prediction by
minimizing the total energy under the lumen-contact constraint, and an $\SE(3)$-equivariant
residual network adds a small Lie-algebra-valued correction composed through the matrix
exponential. The result is projected to 2D and supervised against the angiogram through the optimal-transport loss, with gradients flowing to the network weights and calibration parameters by differentiable simulation. Note that the loss box is labeled Lie-Wasserstein in the schematic; as \S\ref{sec:groundcost} sets out, the loss actually used for training acts on projected positions with a Euclidean ground cost, and the group-intrinsic cost becomes operative only where three-dimensional ground truth exists.}
\label{fig:architecture}
\end{figure}

\subsection{Why this architecture}

\paragraph{Low-data tractability.} The physics simulation handles the bulk of the mapping
deterministically, leaving the network responsible only for systematic deviations between simulator
and reality. Learning a low-dimensional residual on a few dozen cases is plausible where learning
the full nonlinear map is not. Physics-constrained learning has been reported to
reduce data requirements in adjacent fields, including cardiac electrophysiology
\citep{sahlicostabal2020cardiac} and physics-informed fluid modeling
\citep{karniadakis2021piml}. Whether that advantage holds at $n\approx 25$ for this problem is
untested and is the principal risk to the plan of \S\ref{sec:discussion}.

\paragraph{Interpretability.} The physics part is inspectable. The predicted contact-force
distribution, the energy budget at each frame, and the relative contributions of the terms in
Equation~\eqref{eq:Jtotal} are explicit numerical objects with clinical meaning. The residual is
constrained to be small if the physics is right, so a \emph{large} residual at a particular
anatomic region is itself a finding, indicating a physical effect the simulator has under-modeled,
such as adventitial tethering at the visceral segment or hoop stress at a calcified neck. This is
useful for debugging and for communicating with collaborators who are comfortable with mechanics
but not with neural networks.

\paragraph{Equivariance by construction.} The residual network can be implemented with an
$\SE(3)$-equivariant backbone read out invariantly, as the argument above requires, using
group-convolutional networks \citep{cohen2016gcnn},
gauge-equivariant constructions \citep{weiler2021coordinate}, or equivariant attention
\citep{fuchs2020se3transformer}. The constraint means that rotating the patient on the operating table produces a predictably rotated prediction without the network needing to see every rotation in training. Gantry rotation is a separate matter, since the C-arm pose enters only through $\Pi$ and leaves the prediction in patient coordinates unchanged. In the vascular setting specifically, this data
efficiency has been demonstrated directly by \citet{suk2023se3small} and
\citet{suk2024mesh}, whose results motivate the choice here.

\subsection{Training and its difficulties}

The physical parameters split into directly known components, such as wire stiffness from
manufacturer specification and deployment state, and Hounsfield and tortuosity quantities read from
CT, and to-be-calibrated components, principally the anchoring profile and the dependence of $C_v$
on calcification. The latter are calibrated jointly with $\theta$ by minimizing the objective
above.

Gradient flow through the simulator warrants more than a passing remark. The problem is bilevel,
with a complementarity problem at the lower level, and therefore a mathematical program with
equilibrium constraints. The solution map is only piecewise smooth, the derivative jumps when the
active set changes, and standard constraint qualifications fail at complementarity constraints.
Implicit differentiation is valid at a fixed active set \citep{amos2017optnet}, and at active-set
changes the appropriate object is a conservative or Clarke generalized gradient. In practice we
would use the augmented-Lagrangian formulation of \S\ref{sec:kkt} for the backward pass.
This is the mechanism by which the calibration parameters are supposed to be learned, and it is one of the main obstacles between the present work and a trained
model.

One sanity check closes the section. If the physics simulation is perfect, the optimal
residual is zero, the network learns to output zero, the composition with $\Exp$ becomes the
identity, and the prediction equals the simulation. The architecture degrades gracefully toward a
pure physical model when the physical model is good. Conversely, if the physics is badly
misspecified the residual must absorb the error, which the network can in principle do, but at the
cost of needing more data, generalizing less, and losing interpretability.

\section{Numerical Experiments}
\label{sec:numerical}

Before committing to a clinical build, we implemented and exercised the important mathematical
components on synthetic geometries with known ground truth.
First we verify that the kinematics, the contact geometry, the optimal-transport loss and the projection behave as specified, and that the forward model reproduces the qualitative clinical picture, meaning the bowstring effect and apex-concentrated wall loading, on anatomy we constructed and therefore understand exactly.

What follows verifies an \emph{implementation} against
its specification and against geometry we control. It does not validate the model against reality,
because no patient data are used here. Two toy models were built in MATLAB, a planar $\SE(2)$
chord-in-curved-pipe model and a three-dimensional $\SE(3)$ model on a helical iliac. The
Lie-group, optimal-transport and projection primitives pass $18$ of $18$ unit tests, comprising
$6$ hat, vee, exponential, logarithm and group-closure checks on $\SE(3)$, $7$ optimal-transport
primitives, and $5$ projection-geometry checks, and Bishop frames remain orthonormal to machine
precision on a helix. We note plainly that several of these are identities rather than tests. That
$W_2^2(\mu,\mu)=0$, that the transport plan between identical clouds is diagonal, and that an
exponential-logarithm round trip recovers the input are properties of correct code, not evidence
about the physics.

The unit tests cover the Lie-group,
transport and projection primitives, but the constrained mechanics solver itself is verified only
against the qualitative clinical picture and against its own KKT residuals. No component of this
work compares the equilibrium against an independent solution, whether an analytic elastica, a
published benchmark, or a second solver, and the internal consistency checks we do report are derived from the same discretization and so cannot detect an error shared by both sides. We note in particular that the relation $\int_{\mathrm{active}}\lambda\,ds = T_{\mathrm{axial}}\int_{\mathrm{active}}\kappa\,ds$ used in \S\ref{sec:phase2} is not a check at all, since $\lambda$ is \emph{defined} as $T_{\mathrm{axial}}\kappa$; it is a decomposition that explains an agreement, not evidence for it. Independent verification of the mechanics is outstanding.

\subsection{A correction to the mechanics}
\label{sec:correction}

An earlier version of this work included a per-segment Hookean inextensibility spring on the wire,
introduced to suppress node bunching at active-set boundaries. But, as discussed in \S\ref{sec:Jaxial}, this is not right. Its stiffness was three to four orders of magnitude below the wire's
true axial stiffness, and because it was soft it was also the only mechanism permitting the axial
tension to do work. Together the tension term and that spring accounted for $96\%$ of the reported
energy budget, which is to say that most of the energy in the model sat in two mutually
contradictory terms.

The results below use the corrected formulation. The wire slides through the sheath so its
intraluminal arc length is a free degree of freedom, no inextensibility energy appears, and uniform
node spacing is imposed instead as an equality constraint on the parameterization, contributing
nothing to the energy. One implementation subtlety fails silently. Where a boundary condition clamps two adjacent nodes, the segment between them is rigid, and chaining the uniform-spacing equality through that segment freezes the arc length of the whole wire, reproducing exactly the artifact the change was meant to remove. The equality must be applied only to segments having at least one free
endpoint. Under the old formulation the two-dimensional toy reported a
peak vessel displacement of $3.64$~mm. With the artificial spring removed and the wire free to
slide, the same geometry and parameters give $9.97$~mm, which is a far more clinically plausible
magnitude.

\subsection{Phase 1, planar forward model}
\label{sec:phase1}

\paragraph{Methods.} The vessel centerline is a single-period sinusoidal iliac, chosen over a
multi-period sinusoid because real iliacs carry one dominant curve and periodic centerlines admit a
spurious weaving solution, which is itself an instance of the non-convexity noted in
\S\ref{sec:energy}. Centerline and wire are polylines of $N$ nodes lifted to $\SE(2)$ by attaching
tangent-angle frames. The wire is a Lunderquist $0.035''$ guidewire with $r_w=0.4445$~mm. The
equilibrium minimizes $J_{\mathrm{wire}} + J_{\mathrm{vessel}} + J_{\mathrm{anchor}} +
J_{\mathrm{axial}}$ subject to the closest-point lumen inequality of
Equation~\eqref{eq:closestpoint}, enforced hard through a sequential quadratic programming solver
with KKT conditions at a constraint tolerance of $10^{-4}$~mm. Boundary conditions are essential
rather than natural: fixed nodes are eliminated from the design vector and hold their \State{1}
values exactly. Fixing a single end node pins position and leaves the tangent free, and fixing two
adjacent nodes pins position and tangent together, which is what we mean by clamped. Here both wire
ends are single-node pins. We take $s=0$ to be the common femoral access, where the sheath holds
the wire, and the far end to be the parked tip. Because the wire is initialized along the vessel
centerline, which is a feasible start, its two held endpoints lie on the \State{1} centerline by
construction. The axial term then drives the chord-shortening that generates distributed apex
contact. Canonical
parameters are given in Table~\ref{tab:toyparams}.

\begin{table}[t]
\centering
\small
\caption{Canonical parameters for the two toy models. Phase~2 parameters are reported explicitly, since the two models share mechanics but not geometry. Here $EI_v$ denotes the isotropic scalar reduction of the stiffness tensor $C_v(s)$ of \S\ref{sec:Jvessel} used in the toys, and $k_{\mathrm{fat}}$ the uniform toy value of the field $k_{\mathrm{HU}}$ of \S\ref{sec:Jfat}.}
\label{tab:toyparams}
\begin{tabular}{@{}llrrl@{}}
\toprule
\textbf{Parameter} & \textbf{Symbol} & \textbf{Phase 1 (2D)} & \textbf{Phase 2 (3D)} &
\textbf{Rationale} \\
\midrule
Lumen diameter & $D_{\mathrm{lumen}}$ & 12 mm & 12 mm & common iliac luminal diameter \\
Wire outer radius & $r_w$ & 0.4445 mm & 0.4445 mm & Lunderquist $0.035''$ \\
Wire bending modulus & $EI_w$ & 65 N$\cdot$mm$^2$ & 65 N$\cdot$mm$^2$ & swept, see
\S\ref{sec:sweep} \\
Vessel bending modulus & $EI_v$ & 200 N$\cdot$mm$^2$ & 200 N$\cdot$mm$^2$ & assumed \\
Axial pre-tension & $T_{\mathrm{axial}}$ & 10 N & 10 N & assumed surgical pull \\
Anatomic anchoring & $k_{\mathrm{anat}}$ & 0.03 N/mm & 0.03 N/mm & loose tethering, $>0$
required \\
Extravascular & $k_{\mathrm{fat}}$ & --- & 0.04 N/mm & absent in 2D by construction \\
End-to-end chord & $L^{\mathrm{chord}}$ & 150 mm & 150 mm & arc length $185.1$/$177.1$ mm \\
Geometry & --- & sinusoid, $A=25$ mm & helix, $A=15$ mm & one period / one turn \\
Discretization & $N$ & 80 nodes & 60 nodes & see \S\ref{sec:refinement} \\
Boundary condition & --- & pinned-pinned & clamped-pinned & caudal sheath, parked tip \\
\bottomrule
\end{tabular}
\end{table}

\paragraph{Results.} At $N=80$ the solver reaches a first-order optimality of $2.5\times10^{-5}$
with a maximum constraint violation of $6.5\times10^{-13}$~mm after $905$ iterations. We report
these rather than the solver's exit flag, because the flag returned indicates that the step size
fell below tolerance rather than that optimality was certified, and quoting it as ``converged
cleanly'' would be misleading.

\begin{figure}[t]
\centering
\includegraphics[width=\textwidth]{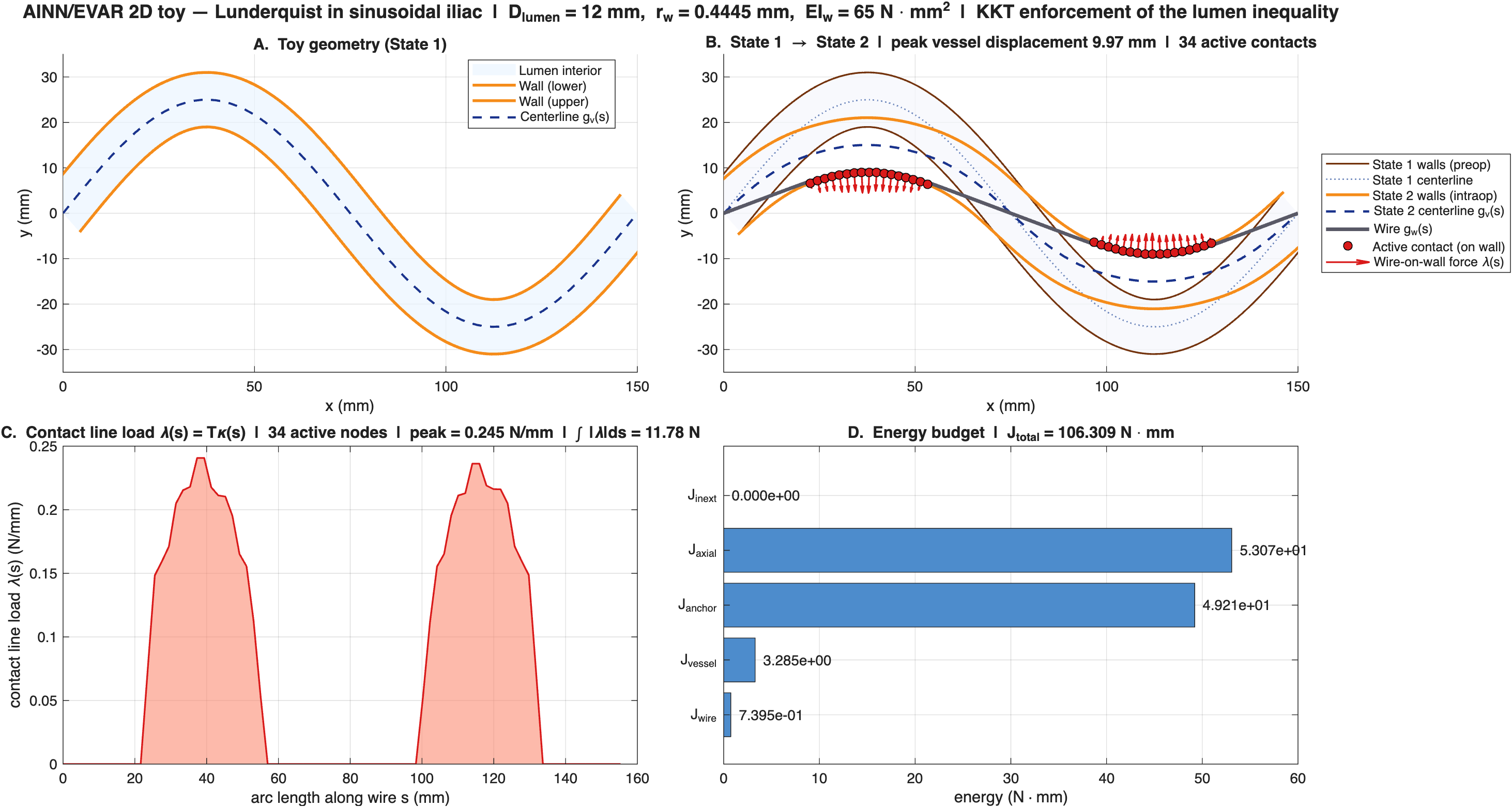}
\caption{\textbf{Phase 1 planar toy at equilibrium.} (A) \State{1} geometry, lumen wall and centerline. (B) \State{1} $\rightarrow$ \State{2} overlay with the equilibrium wire and the active wall contacts. (C) Contact line load $\lambda(s)=T\kappa(s)$ over the active set. The shaded trace is median-filtered over three nodes for display, because a single node's discrete curvature can spike at an active-set boundary where the local Voronoi length collapses; the peak and integral in the panel title are the raw solver values. (D) Energy budget. The inextensibility term is identically zero by construction (\S\ref{sec:correction}).}
\label{fig:phase1}
\end{figure}

The equilibrium reproduces the canonical picture (Figure~\ref{fig:phase1}). The wire bowstrings inside the lumen, taking the chord across each curve, its arc length shortening from the $185.1$~mm of the undeformed sinusoidal path toward the $150$~mm chord and reaching $155.3$~mm, and the vessel deforms with a peak centerline displacement of $9.97$~mm and a mean of $5.43$~mm. Thirty-four of the $80$ wire nodes are in active contact with
the wall, in two clusters, one at each curvature apex. The contact reaction, reported as the line load $\lambda(s)=T_{\mathrm{axial}}\kappa(s)$ restricted to the active set, is smooth and center-peaked at the apexes with a peak of $0.24$~N/mm and an integrated wall reaction of $11.78$~N. The energy budget is dominated by the axial term at $53.1$~N$\cdot$mm, followed by
anchoring at $49.2$, vessel bending at $3.3$, and wire bending at only $0.74$~N$\cdot$mm, giving a
total of $106.3$~N$\cdot$mm.

\subsection{Phase 2, three-dimensional forward model}
\label{sec:phase2}

\paragraph{Methods.} Phase 2 lifts the identical mechanics to $\SE(3)$, with closed-form Rodrigues
exponential and logarithm maps and a Bishop parallel-transport moving frame, so framing stays
smooth on near-straight segments. The synthetic vessel is a helical iliac of one full turn over the
segment length. The energy gains the Winkler extravascular term $J_{\mathrm{fat}}$ with uniform toy
density, which is intentionally absent in two dimensions. The boundary condition is
clamped-pinned, the clamp realized by fixing the two nodes adjacent to the femoral access and the
pin by fixing the single node at the parked cranial tip. Wire bending is
curvature-only, which as noted in \S\ref{sec:Jwire} is a consequence of the Bishop framing rather
than an independent physical assumption.

\begin{figure}[t]
\centering
\includegraphics[width=\textwidth]{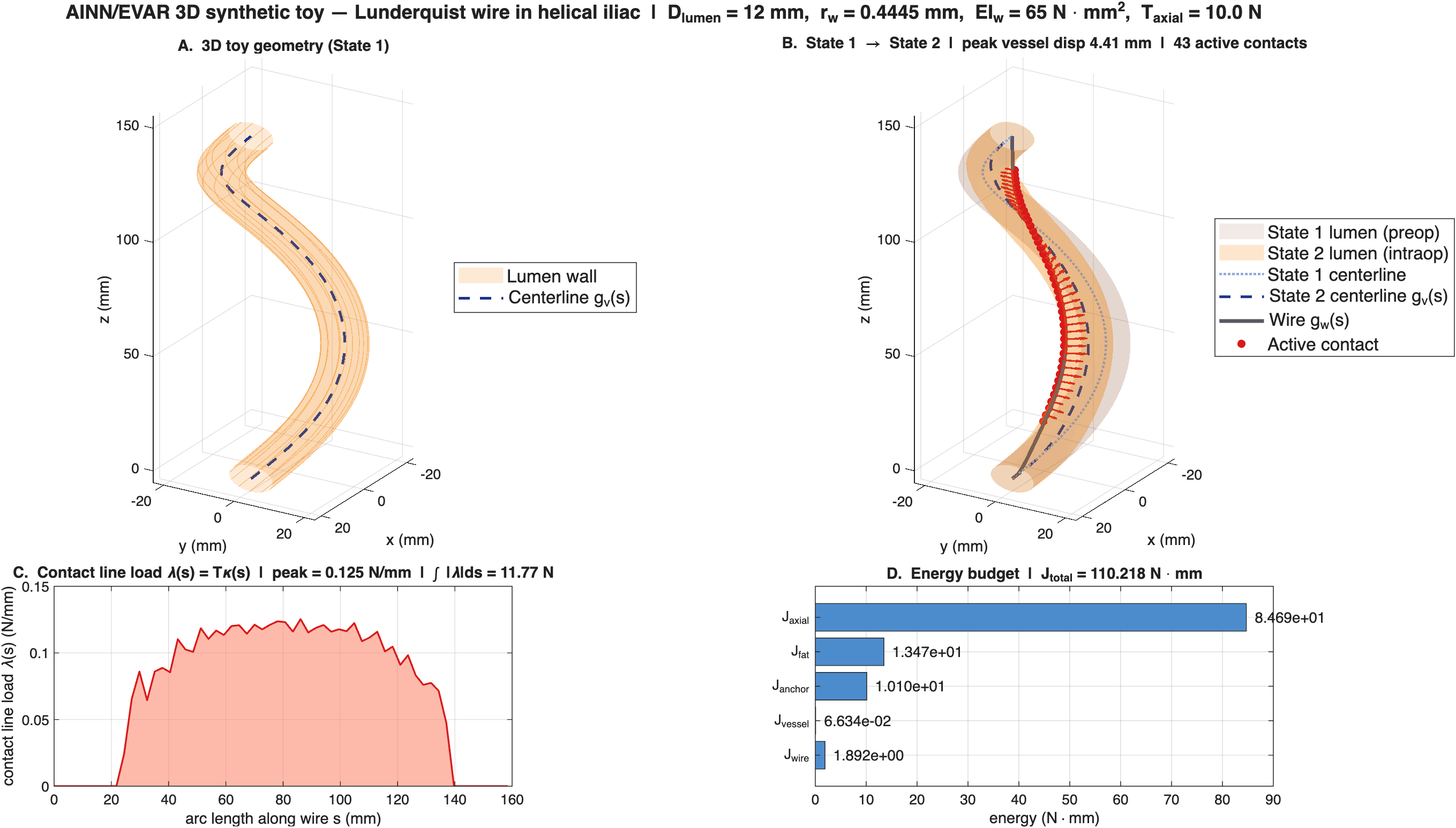}
\caption{\textbf{Phase 2 three-dimensional synthetic toy at equilibrium}, a Lunderquist wire in a helical iliac. (A) \State{1} geometry, lumen-wall tube and centerline. (B) \State{1} $\rightarrow$ \State{2} overlay with equilibrium wire and active contacts, the \State{1} lumen translucent behind the deformed \State{2} lumen. (C) Contact line load along arc length. (D) Energy budget.}
\label{fig:phase2}
\end{figure}

\paragraph{Results.} The solver reaches a first-order optimality of $5.5\times10^{-5}$ with maximum
constraint violation $4.6\times10^{-11}$~mm after $290$ iterations. The bowstring reproduces in three dimensions (Figure~\ref{fig:phase2}), with contact distributed as a single extended patch along the helical turn rather than as the two localized apex patches of the planar toy, which is what the geometry predicts. Peak vessel displacement is $4.41$~mm, mean $2.90$~mm, with $43$ of $60$ nodes in active contact and a line load peaking at $0.125$~N/mm. The integrated wall reaction is $11.77$~N, against $11.78$~N in two dimensions. That near-equality is not coincidental. Since $\int_{\mathrm{active}}\lambda\,ds = T_{\mathrm{axial}}\int_{\mathrm{active}}\kappa\,ds$, it is the tension multiplied by the total turning angle over the contact set, and the two toys happen to turn through nearly the same angle in contact. The wire's arc length
shortens from $177.1$~mm toward the $150$~mm chord, reaching $158.5$~mm. As in two dimensions the
axial term dominates the budget at $84.7$ of $110.2$~N$\cdot$mm, with extravascular resistance
$13.5$, anchoring $10.1$, wire bending $1.9$ and vessel bending $0.07$~N$\cdot$mm. The smaller
displacement relative to two dimensions reflects the added extravascular resistance and the stiffer
clamped boundary, both physically expected. The lift $\R^3\to\SE(3)\to\R^3$ recovers the original
centerline to numerical tolerance, which confirms the Rodrigues implementation.

\subsection{Sensitivity to wire bending stiffness}
\label{sec:sweep}

\begin{figure}[t]
\centering
\includegraphics[width=\textwidth]{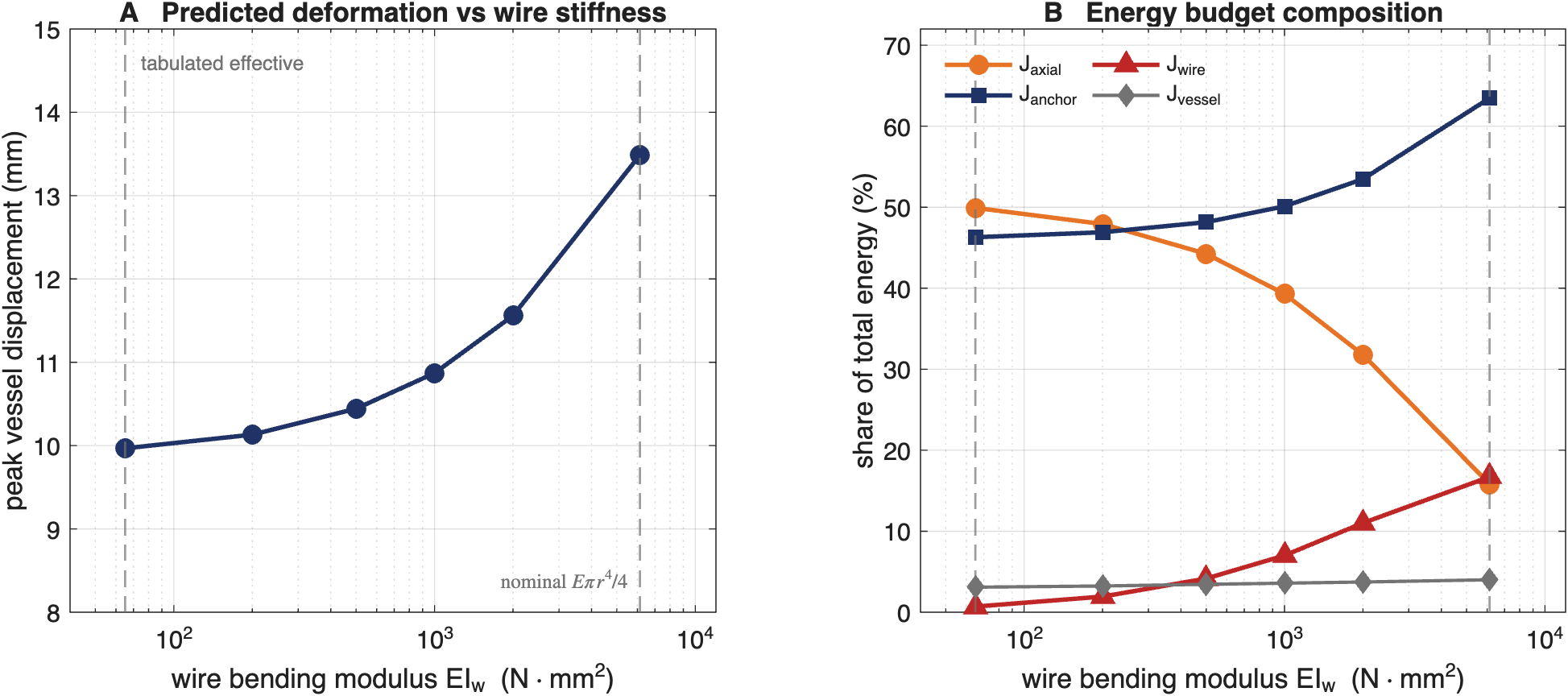}
\caption{\textbf{Sensitivity of the planar (Phase 1) equilibrium to wire bending stiffness.} The swept interval spans the effective stiffness commonly assumed for a Lunderquist through the nominal solid-rod value implied by $E\pi r_w^4/4$. (A) Peak vessel displacement. (B) Composition of the energy budget.}
\label{fig:sweep}
\end{figure}

Because the effective bending stiffness of a clinical guidewire is uncertain by roughly two orders
of magnitude (\S\ref{sec:Jwire}), we swept $EI_w$ across the entire plausible interval, from $65$
to the nominal solid-stainless value $E\pi r_w^4/4 = 6132$~N$\cdot$mm$^2$, holding all other
parameters fixed (Table~\ref{tab:sweep}, Figure~\ref{fig:sweep}). The planar toy was swept at six values and the three-dimensional toy at the two endpoints of the interval.

\begin{table}[t]
\centering
\small
\caption{Wire-stiffness sweep. Energy shares are percentages of $J_{\mathrm{total}}$. The
extravascular term is absent from the planar toy by construction.}
\label{tab:sweep}
\begin{tabular}{@{}rrrrrrrr@{}}
\toprule
$EI_w$ (N$\cdot$mm$^2$) & peak (mm) & mean (mm) & $J_{\mathrm{wire}}$ &
$J_{\mathrm{axial}}$ & $J_{\mathrm{anchor}}$ & $J_{\mathrm{fat}}$ & $J_{\mathrm{vessel}}$ \\
\midrule
\multicolumn{8}{@{}l@{}}{\emph{Phase 1, planar, $N=80$}} \\[2pt]
65    & 9.97  & 5.43 & 0.7\%  & 49.9\% & 46.3\% & --- & 3.1\% \\
200   & 10.13 & 5.49 & 1.9\%  & 47.9\% & 46.9\% & --- & 3.2\% \\
500   & 10.44 & 5.63 & 4.2\%  & 44.2\% & 48.1\% & --- & 3.4\% \\
1000  & 10.87 & 5.85 & 7.0\%  & 39.3\% & 50.1\% & --- & 3.6\% \\
2000  & 11.56 & 6.24 & 11.0\% & 31.8\% & 53.5\% & --- & 3.7\% \\
6132  & 13.49 & 7.41 & 16.7\% & 15.8\% & 63.5\% & --- & 4.0\% \\
\midrule
\multicolumn{8}{@{}l@{}}{\emph{Phase 2, three-dimensional, $N=60$}} \\[2pt]
65    & 4.41  & 2.90 & 1.7\%  & 76.8\% & 9.2\%  & 12.2\% & $<0.1\%$ \\
6132  & 6.33  & 3.65 & 36.5\% & 42.0\% & 9.2\%  & 12.2\% & $<0.1\%$ \\
\bottomrule
\end{tabular}
\end{table}

Here, each stiffness level was warm-started from the previous one. The intermediate levels required between
$522$ and $905$ iterations, but the final level at $EI_w=6132$ terminated after a single reported iteration, having moved peak displacement from the $11.56$~mm of the $EI_w=2000$ warm start to $13.49$~mm. A single step that large is not on its own credible, and because that row supplies the upper endpoint of the sensitivity claim, we re-solved it independently from a cold start on the undeformed centerline. The cold solve took $1122$ iterations and reproduced the
warm-started row to five significant figures, giving peak displacement $13.4868$ against
$13.4869$~mm and total energy $143.5420$ against $143.5420$~N$\cdot$mm, with first-order optimality
improved from $7.0\times10^{-3}$ to $2.7\times10^{-4}$ and constraint violation from
$3.4\times10^{-7}$ to $8.8\times10^{-11}$~mm. The endpoint is therefore path-independent within the
precision we report, though this is a restart check rather than verification against an independent
solver.

Two conclusions follow. First, \textbf{axial pre-tension rather than wire bending supplies the dominant
energetic drive}. We find this surprising and not clinically intuitive, and we are not certain whether that reflects an incomplete physical picture of what the axial pre-tension represents, or something about the formulation that does not correspond to the procedure.
The finding is that at the low end of the range wire bending is under one percent of the budget, and
even at the nominal solid-rod stiffness the two are merely comparable, with bending edging ahead at $16.7\%$ against $15.8\%$, rather than either dominating. This ordering holds at every mesh resolution we tested (\S\ref{sec:refinement}), where
the wire-bending share stays below one percent at $EI_w=65$ from $N=40$ through $N=160$.

Second, and more usefully, \textbf{the predicted deformation is insensitive to $EI_w$}. A $94\times$ change in wire stiffness moves peak displacement by only $35\%$ in two dimensions and $44\%$ in three, the latter from $4.41$ to $6.33$~mm at the two endpoints of the range (Table~\ref{tab:sweep}). The sweep was run at the canonical discretization only, so we claim the insensitivity qualitatively and do not assert that the ratio itself is mesh-converged. This matters practically. The uncertainty in guidewire
effective stiffness, which we could not resolve from published sources, turns out to be largely
irrelevant to the prediction, because the deformation is set by the axial load and the vessel's
restoring stiffness rather than by the wire's resistance to bending.

\subsection{Mesh refinement, and a negative result}
\label{sec:refinement}

Because the discrete strain of Equation~\eqref{eq:discretestrain} carries an explicit node-spacing
factor, mesh dependence must be checked rather than assumed. We refined the Phase 1 toy from $N=40$
to $N=160$, warm-starting each level by arc-length interpolation from the previous
(Table~\ref{tab:refine}).

\begin{table}[t]
\centering
\small
\caption{Phase 1 mesh refinement at $EI_w=65$~N$\cdot$mm$^2$. Energies in N$\cdot$mm.}
\label{tab:refine}
\begin{tabular}{@{}rrrrrrrr@{}}
\toprule
$N$ & peak disp. (mm) & $L^{\mathrm{arc}}_w$ (mm) & $J_{\mathrm{wire}}$ & $J_{\mathrm{vessel}}$ &
$J_{\mathrm{anchor}}$ & $J_{\mathrm{axial}}$ & $J_{\mathrm{wire}}$ share \\
\midrule
40  & 13.10 & 152.41 & 0.350 & 3.671 & 44.27 & 24.10 & 0.48\% \\
60  & 11.30 & 153.93 & 0.553 & 3.449 & 48.62 & 39.27 & 0.60\% \\
80  & 9.97  & 155.31 & 0.739 & 3.285 & 49.21 & 53.07 & 0.70\% \\
120 & 8.10  & 157.54 & 1.023 & 2.842 & 46.94 & 75.40 & 0.81\% \\
160 & 6.89  & 159.18 & 1.231 & 2.474 & 43.83 & 91.84 & 0.88\% \\
\bottomrule
\end{tabular}
\end{table}

\textbf{The predicted displacement magnitude
is not mesh-converged.} Peak displacement falls monotonically from $13.10$ to $6.89$~mm as $N$
quadruples, and the decrement per doubling of $N$ is close to constant at roughly $3$~mm, which is
the signature of logarithmic rather than convergent behavior. The equilibrium wire arc length rises
correspondingly, from $152.4$ to $159.2$~mm.

The lumen inequality is imposed at nodes, so a coarse
discretization permits the wire to cut corners between constraint points, shorten more than it
physically could, and therefore push harder on the wall. Refinement progressively removes that
freedom. A consistent discretization would impose non-penetration segment-wise rather than
node-wise, and we expect that to restore convergence; implementing it is the most important
numerical item of future work. We also note that convergence of the solver itself degrades with
$N$, with first-order optimality reaching only $7\times10^{-1}$ at $N=160$ against
$2.5\times10^{-5}$ at $N=80$, so part of the drift at the finest level may be incomplete
convergence rather than discretization.

The qualitative picture, the
bowstring, the two apex-localized contact patches, and the fraction of nodes in contact at roughly
$40\%$ in this planar toy, is however stable across the whole range. The ordering of the energy terms, with
wire bending under one percent at $EI_w=65$, is stable. And the insensitivity to $EI_w$ is stable. The absolute displacement
is not, and any comparison of this model against clinical measurements must wait for the
segment-wise contact discretization. We therefore make no quantitative claim about predicted
millimeters of ostial migration.

Two further limits of this study should be stated. First,
the trend over the range tested does not identify a limit. Extrapolating the observed decrement
naively would drive the predicted displacement toward zero within a further two or three
refinements, which is not a physically sensible answer, and we do not know whether the true
behavior is a slow approach to a nonzero value or a genuine collapse of the node-wise formulation.
Distinguishing the two requires the consistent discretization, not more levels of the inconsistent
one. Second, the sweep of \S\ref{sec:sweep} varies $EI_w$ alone. $T_{\mathrm{axial}}$ is fixed at
the assumed $10$~N throughout and is never varied, even though the energy budget identifies it as
the dominant term and $J_{\mathrm{axial}}$ is linear in it, so the equilibrium is certainly more
sensitive to $T_{\mathrm{axial}}$ than to the parameter we did sweep. The insensitivity result of
\S\ref{sec:sweep} is therefore a statement about wire stiffness specifically and must not be read
as robustness of the model as a whole.

\subsection{Optimal-transport loss}
\label{sec:otvalidation}

\begin{figure}[t]
\centering
\includegraphics[width=\textwidth]{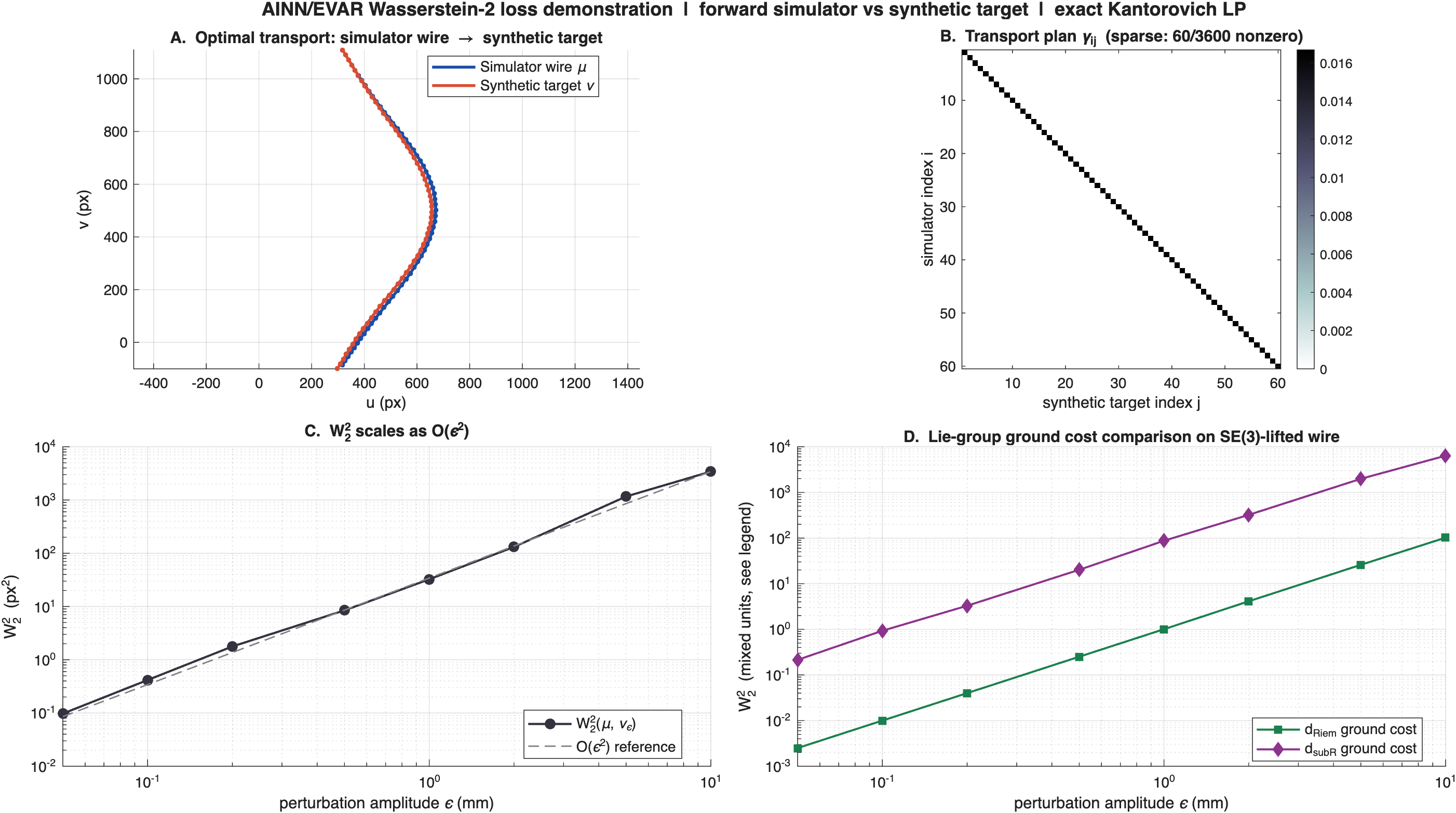}
\caption{\textbf{Optimal-transport loss on the $\SE(3)$-lifted wire.} (A) Transport between the simulator wire measure and a synthetic target. (B) Transport plan. (C) $W_2^2$ against perturbation amplitude. (D) Nilpotent sub-Riemannian ground cost against the Riemannian cost.}
\label{fig:wasserstein}
\end{figure}

The Wasserstein-2 data term of \S\ref{sec:wasserstein} is computed by the exact Kantorovich linear
program, with a vectorized squared-Euclidean fast path and support for custom ground costs
(Figure~\ref{fig:wasserstein}). Three expected properties were checked on $\SE(3)$-lifted wire
measures. The self-distance vanishes. For a rigid perturbation of amplitude $\varepsilon$ the loss
scales as $O(\varepsilon^2)$, confirmed across more than two decades of $\varepsilon$. And for matched
discretizations the optimal plan is the sparse diagonal coupling, with $60$ of $3600$ entries
nonzero, so the LP recovers the identity matching when source and target coincide.

As noted above, these are consistency checks rather than tests of the physics, and the
$O(\varepsilon^2)$ scaling under a rigid translation is exact by definition rather than an
empirical finding. The property that actually matters for training, namely the behavior of the loss
at transport-plan switches where it is non-smooth, is not probed by a rigid translation and remains
to be characterized. The same panel compares the Riemannian ground cost against the nilpotent
first-order approximation to the sub-Riemannian cost on the lifted wire. The sub-Riemannian cost
assigns systematically higher cost to lateral motion, which is the non-holonomic penalty motivating
the choice in \S\ref{sec:groundcost}.

\subsection{Single-view projection}
\label{sec:projvalidation}

\begin{figure}[t]
\centering
\includegraphics[width=\textwidth]{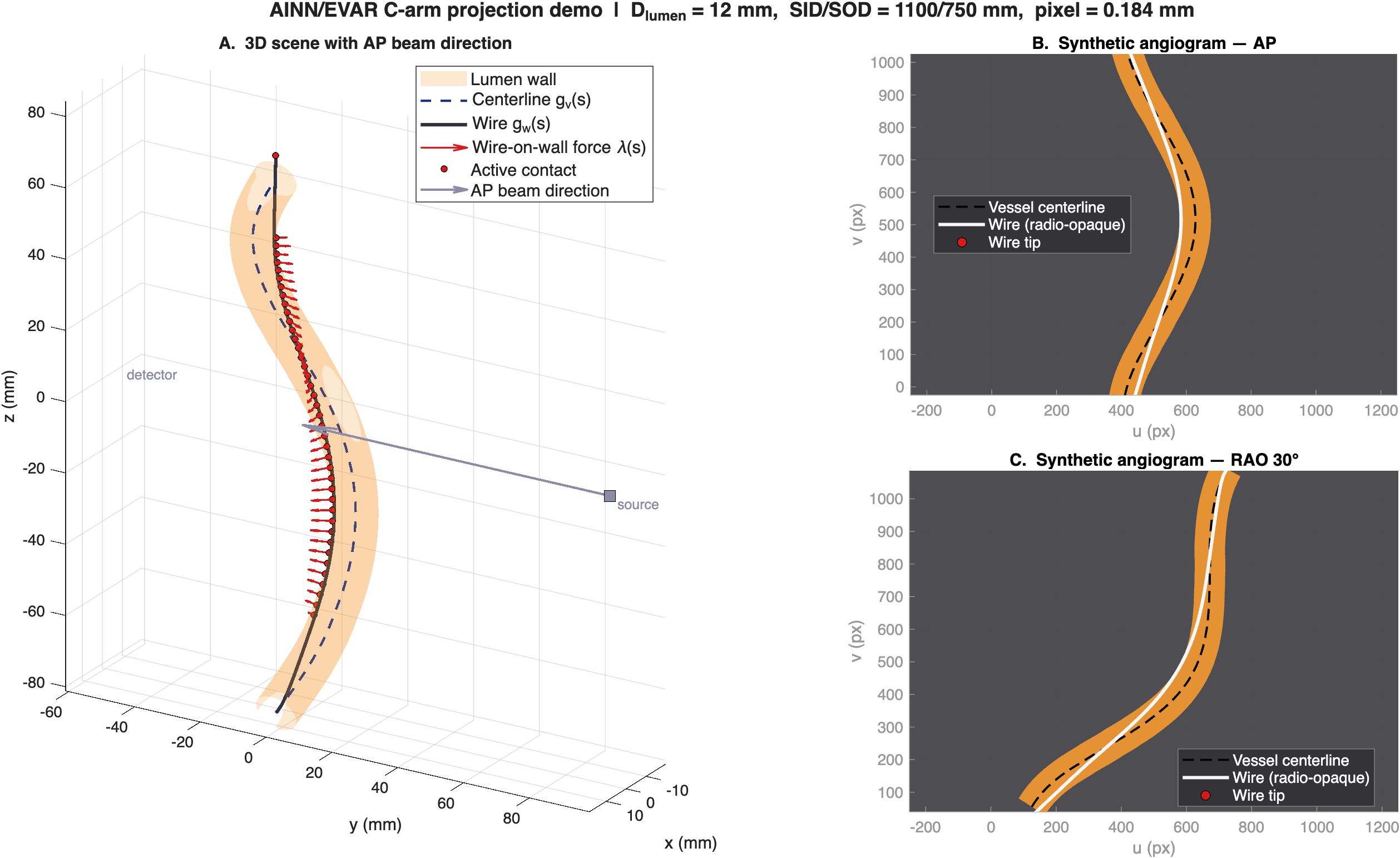}
\caption{\textbf{C-arm projection.} (A) Three-dimensional scene with lumen wall, centerline,
equilibrium wire, wall-contact forces, and the beam axis. (B) Synthetic anteroposterior angiogram
and (C) synthetic $30^\circ$ oblique angiogram of the Phase 2 toy, with the wire rendered
radio-opaque against the projected centerline.}
\label{fig:projection}
\end{figure}

The supervision pathway of \S\ref{sec:projection} requires projecting the predicted
three-dimensional configuration to a detector plane matching the intraoperative angiogram. The
C-arm pose is built from gantry primary and secondary angles together with source-to-image and
source-to-object distances, following the standard oblique and cranio-caudal convention, and the
projection is a perspective point-source map into detector millimeters or pixels. Three geometric
invariants were verified. A landmark at the isocenter maps to the detector principal point, the
transverse magnification equals the ratio of source-to-image to source-to-object distance, and a
$90^\circ$ oblique rotation produces the expected image-axis swap. Figure~\ref{fig:projection}
shows the three-dimensional scene together with two synthetic angiograms of the Phase 2 toy, an
anteroposterior view and a $30^\circ$ oblique view, with the wire rendered radio-opaque against the
projected vessel centerline, demonstrating the complete three-dimensional-prediction to
two-dimensional-observation path on which the supervision of \S\ref{sec:projection} depends.

\subsection{What the toy program establishes, and what it does not}
\label{sec:whatestablished}

The two toys verify, on data whose ground truth we control, the $\SE(2)$ and $\SE(3)$ kinematics and exponential-map updates, the closest-point unilateral lumen contact, the exact Wasserstein-2 loss with both Riemannian and approximate sub-Riemannian ground costs, and the perspective C-arm projection. The anchored energy functional and its KKT solution are exercised but, as \S\ref{sec:numerical} states, checked only against their own optimality conditions.

Beyond verification, the program produced two modeling decisions the production pipeline inherits and that are not forced by the formulation. The anatomic anchoring stiffness must be strictly positive everywhere to break the degenerate phase-shift solution, and toy geometry should carry a single dominant curve, matching real iliac anatomy and avoiding a multi-zero-crossing weaving artifact.

The limitations are equally concrete. \S\ref{sec:limitations} collects them; two belong here because they bear directly on how these results should be read.
\begin{itemize}[leftmargin=1.6em,itemsep=2pt,topsep=3pt]
\item \textbf{The problem is non-convex} and the solver returns local minima. Two distinct spurious
minima were encountered and worked around by geometry and parameter choices rather than eliminated.
\item \textbf{The bowstring is not evidence.} The toys were constructed so that $r_w < \Rlum$, and
the bowstring follows from that inequality alone. Reproducing it confirms the implementation, not
the model's fidelity to a patient.
\end{itemize}

\section{Discussion}
\label{sec:discussion}

This work introduces AINN and applies it to a real surgical problem that surgeons currently resolve by intuition, and that the next generation of endovascular robotics will instead need to compute. Whether a given aorta will straighten enough to compromise a landing zone or displace a branch ostium is, today, a judgment. For an autonomous agent it has to become a prediction.

Prior research by others on this problem was informative. Representing vessel centerlines as a wireframe of frames in $\SE(3)$, and applying a Wasserstein loss in this domain, as well as the three-dimensional-to-two-dimensional state representation, which is standard practice wherever a projection supervises a volumetric prediction, have each been used before. We, however, hook these pieces together into a single coupled formulation, in which the same group carries the state, the contact geometry, the residual correction and the projection. Second, and more generally, the paradigm-level contribution is not that anatomy can enter a loss at all, since shape, topology and inequality priors already do that (\S\ref{sec:relatedwork}), but the pairing of \emph{soft anatomic priors in the loss with hard anatomic priors carried by the state representation}, made an explicit per-prior design choice and organized into categories, of which the vascular problem here is just a worked example.

Within the worked example, the piece that appears novel is the anatomically structured mobility and anchoring field of \S\ref{sec:Janchor}. A position-dependent stiffness $k_{\mathrm{anat}}(s)$, encoding as a population-level prior the surgical fact that the arch is fixed and the external iliacs are mobile, does not appear as a prior category in the existing surveys of domain knowledge in medical imaging (\S\ref{sec:taxonomy}), and in the planar toy's energy budget it carries more of the deformation than any term other than the axial load, though in the three-dimensional toy the extravascular term outranks it.

\subsection{What the representation buys}

Finite-element treatments of this problem are more mature than what is presented here, and they are
validated on patients, which this is not. What the $\SE(3)$ formulation offers is not better
mechanics but a smaller problem, posed in the coordinates the case is planned in. The Phase 1 and Phase 2 models carry a few hundred degrees of freedom and solve in $30$ and $11$ seconds respectively on a laptop. That is comfortable for offline study but too slow for a training loop, as \S\ref{sec:limitations} sets out; warm-started incremental solves, or a learned surrogate for the equilibrium map, are the obvious routes and neither is exercised here.

\subsection{Why AINN is naturally loss-based}

The clinical stakes scale with autonomy and anatomic complexity, and the anatomy itself explains why AINN is naturally loss-based rather than constraint-based. Physics-informed networks are
sometimes faulted because their governing equations enter only as a soft penalty, so a network can
in principle learn a merely expensive rather than a physically impossible solution. For anatomy this is a feature rather than a defect, for the reason set out in \S\ref{sec:hardsoft}: encoding anatomic structure as a distribution rather than as a manifold is the natural choice for biology, because a hard constraint would forbid configurations that are rare but real.

\subsection{Limitations}
\label{sec:limitations}

We have found an approach that appears to
work on geometry we control, and that is a different thing from having found the best approach. At
almost every choice below there are alternatives we have not tried, and without a library of real
preoperative CTs to adjudicate between them we cannot claim the choice we made is the right one.

\paragraph{Scope.} No patient data are used anywhere in this work, and nothing here is validated
against a measurement. \S\ref{sec:numerical} verifies an implementation against synthetic geometry we
constructed. The residual network of
\S\ref{sec:architecture} is specified but not yet trained.

\paragraph{Numerics.} The most immediate is the mesh-convergence failure of \S\ref{sec:refinement}, which must be resolved before any displacement magnitude from this model can be compared with a measurement. Beyond that, whether a constrained interior-point solve suffices, or a structure-preserving Lie-group
integrator is needed for stability under stiff wire-and-device assemblies, is best settled
empirically. We say structure-preserving rather than symplectic deliberately, since
\S\ref{sec:energy} establishes that the constrained system is not Hamiltonian. And
differentiating through an active-set solve is a mathematical program with equilibrium constraints
(\S\ref{sec:architecture}), which is the principal technical obstacle to training the residual at
all. Three further numerical limits belong here. The problem is non-convex and the solver returns local minima, with two spurious minima worked around by geometry and parameter choices rather than eliminated (\S\ref{sec:whatestablished}). The mechanics solver has never been checked against an independent solution (\S\ref{sec:numerical}). And the equilibrium solves, though comfortable offline, are too slow to sit inside a training loop, since a single epoch over a few dozen cases would require thousands of forward solves.

\paragraph{Modeling.} The anchoring profile and the wire-and-device effective stiffness are calibration targets to be fit against data and cross-checked with published vessel-mobility measurements, not quantities to be assumed. This is the flip side of the anchoring field being a contribution. The rigid-wall assumption excludes the large-sheath regime (\S\ref{sec:assembly}). Neither toy contains a bifurcation, so the topological priors the paradigm advertises are
not yet exercised by the worked example. And the axial pre-tension that dominates the energy budget is an assumption awaiting force-sensor calibration, and was never swept (\S\ref{sec:refinement}). Contact is also assumed frictionless (\S\ref{sec:Jaxial}).

\paragraph{Architecture.} The residual composition of \S\ref{sec:architecture} acts per frame, so until one of the two remedies given there is implemented, the arc-length and contact-feasibility guarantees of the forward model apply to the simulator output and not to the final prediction.

\paragraph{Anatomic containment.} The envelope constraint of \S\ref{sec:envelope} is specified but
not implemented, so in the models actually run nothing prevents a displaced centerline from leaving
the patient except the soft anchoring and extravascular terms.

\paragraph{Loss geometry.} The optimal-transport ground cost admits a pragmatic progression from a
Riemannian approximation to the exact sub-Riemannian distance, and the manner in which a
centerline-and-radius profile is converted into a measure, whether uniform along arc length,
weighted by lumen cross-section, or weighted toward landing zones and ostia, shapes what the loss
rewards. None of this is exercised by a projected 2D loss, which also leaves the residual's rotational degrees of freedom only weakly identified (\S\ref{sec:groundcost}).

\paragraph{Data.} The three-dimensional preprocessing required to instantiate this on patients, meaning lumen segmentation, centerline extraction and frame attachment, is treated separately and is out of scope here. The two-dimensional side, meaning wire extraction and
landmark annotation from the angiogram, requires bespoke but bounded effort, and the rigid registration between the preoperative CT frame and the C-arm frame is the single largest source of systematic error and warrants dedicated validation. If that registration is fit against the same landmarks that then appear in the landmark term of the loss, the term is partially self-satisfied, so it must be fit on disjoint features, for instance bony landmarks, or the landmark term excluded from the evaluation metric.

\subsection{What a real evaluation would look like}

Definitive evaluation awaits a clinical cohort. A realistic target is on the order of twenty-five paired
preoperative-CT and intraoperative angiogram cases from a single high-volume center, assessed by
two-dimensional Hausdorff and Wasserstein distances between predicted and observed wire paths and
landmarks, with a held-out subset carrying completion cone-beam CT for absolute three-dimensional
error, under leave-one-out cross-validation appropriate to the cohort size. For that validation to
mean anything, the entire calibration, including the anchoring profile and the stiffness
coefficients, must be refit inside each fold rather than fit once on the full cohort, otherwise the
cross-validation is optimistically biased.

\citet{mohammadi2018planning} report
displacement errors of roughly $3$~mm at iliac ostia against intraoperative fluoroscopy, and
\citet{emendi2023guidewire} report in silico against in vitro agreement of $1.2$ to $1.4$~mm. A
method that cannot approach those numbers has not earned its added machinery, whatever its
representational advantages. Stating the target in advance seems to us the minimum honest posture
for a preprint that proposes a new formulation for an already-studied problem.

\section{Conclusion}
\label{sec:conclusion}

We introduced Anatomy-Informed Neural Networks, a paradigm that encodes anatomic structure directly into a deep-learning model, both as soft priors in the loss and as hard priors built into the architecture and state representation. We organized anatomic priors into categories spanning topology,
mobility, symmetry, shape, atlas-relative position, and contact, taking care to situate that
organization against the existing surveys of prior knowledge in medical imaging, and identifying
mobility and anchoring as the category we believe is genuinely underserved.

We developed the paradigm on a clinically consequential test case, predicting how the aortic and
iliac tree deforms when a stiff guidewire is introduced. Our departure from the established finite-element literature is representational: we take the centerline and its moving frame as the primitive object, which places the state in the Lie group $\SE(3)$. From that choice, a coupled Cosserat-rod formulation with a single unilateral lumen inequality follows
directly, together with an optimal-transport loss between measures of frames and a projection
operator that lets a routine single-view angiogram supervise a three-dimensional prediction.

We verified the forward mechanics, the loss and the projection in silico on synthetic problems with
known ground truth. In the course of doing so we corrected a defect in an earlier formulation (\S\ref{sec:correction}). Removing it raises the predicted peak displacement in the planar toy from $3.64$ to $9.97$~mm and leaves an energy budget containing only physical terms. A sweep across the full two-order-of-magnitude uncertainty in guidewire bending
stiffness shows that axial pre-tension rather than wire bending supplies the dominant energetic drive across the clinically plausible range, the two becoming comparable only at the nominal solid-rod stiffness, and, more usefully, that the predicted deformation is largely insensitive to that uncertain
parameter. A mesh-refinement study shows that the qualitative behavior and the energy ordering are
stable but the absolute displacement magnitude is not yet converged, which we report as a negative
result and as the most pressing item of numerical future work.

What is established here is a formulation and its in silico verification. Whether
the representational argument translates into better predictions than the meshed finite-element
models that came before is an empirical question, and the existing literature has already set the
numbers to beat. AINN is offered as a general framework for anatomically constrained prediction;
its generality remains a claim to be tested on further anatomic domains, and the vascular worked
example is the first of them.

\section*{Scope and Status of This Preprint}
\addcontentsline{toc}{section}{Scope and Status of This Preprint}

This preprint presents a mathematical formulation together with its in silico verification. To avoid any ambiguity about what is and is not established here, we state it plainly. No neural network was trained, no patient data were used, no clinical images appear, and no predictive accuracy on patients is claimed or implied. The architecture of \S\ref{sec:architecture} is a specification.

\section*{Data and Code Availability}
\addcontentsline{toc}{section}{Data and Code Availability}

The two toy models, the Lie-group, optimal-transport and projection primitives, and the scripts
generating Figures~\ref{fig:phase1}--\ref{fig:projection} are implemented in MATLAB and are
available from the author on reasonable request. The toy geometries are fully specified by Table~\ref{tab:toyparams}.

\paragraph{Software and hardware.} All results were produced in MATLAB R2025b
(\texttt{25.2.0.3150157}, Update 4) using the Optimization Toolbox, whose \texttt{fmincon}
sequential quadratic programming solver computes the constrained equilibria of
\S\ref{sec:numerical} and whose \texttt{linprog} simplex solver computes the exact Kantorovich
program of \S\ref{sec:wasserstein}. No other toolbox is required. All runs were performed on a
single MacBook Pro (13-inch, M1, 2020) with an Apple M1 processor, eight cores, $16$~GB of unified
memory, under macOS $26.5$, in serial and without GPU acceleration. The solve times quoted in
\S\ref{sec:discussion} refer to this machine.

\section*{Declaration of Generative AI Use}
\addcontentsline{toc}{section}{Declaration of Generative AI Use}

During the preparation of this work the author used Anthropic's Claude to assist with development of the schematic figures (Figures~\ref{fig:concept} through \ref{fig:architecture}), with an adversarial technical audit of an earlier draft, with document formatting, and with editing of the text for clarity and consistency. Figure~\ref{fig:aortoiliac} was subsequently corrected and redrawn by the author, whose anatomic and mechanical judgment determines its final content. The numerical result figures, Figures~\ref{fig:phase1} through \ref{fig:projection}, were produced by the author's own MATLAB simulation code. The scientific content, the mathematical formulation, and all claims are
the author's own. The author reviewed and edited the generated output and takes full
responsibility for the content of this manuscript.

\section*{Ethics}
\addcontentsline{toc}{section}{Ethics}

This preprint reports mathematical development and in silico verification only. No human subjects,
animal subjects, or patient data were involved, and all geometries presented are synthetic. The
broader research program from which this work derives is approved by the Johns Hopkins Medicine
Institutional Review Board (IRB00562871).

\section*{Funding and Competing Interests}
\addcontentsline{toc}{section}{Funding and Competing Interests}

No funding was used for this work. The author declares no competing interests.

\phantomsection
\addcontentsline{toc}{section}{References}
\bibliographystyle{unsrtnat}
\bibliography{refs}

\end{document}